\documentclass[]{fairmeta/fairmeta}
\usepackage[dvipsnames]{xcolor}
\usepackage{colortbl}
\usepackage{bbding}
\usepackage{booktabs}
\usepackage{caption}
\usepackage{enumitem}
\usepackage{graphicx}
\usepackage{makecell}
\usepackage{mathpazo}
\usepackage{mathtools}
\usepackage{pgfplots}
\usepackage{subcaption}
\usepackage{tgpagella}
\usepackage{tikz}
\usepackage{wrapfig}
\usepackage{xfrac}
\usepackage{xspace}
\definecolor{methodpurple}{RGB}{76,40,130}
\usetikzlibrary{spy}
\newcommand{\mname}{OverSeer\xspace}
\newcommand{\benchname}{MAC-Bench\xspace}

\definecolor{citeblue}{rgb}{0.16, 0.24, 0.65}
\definecolor{benchblue}{RGB}{224,235,247}
\definecolor{taskblue}{RGB}{240,245,250}
\definecolor{metricgray}{RGB}{246,246,246}
\definecolor{oursgreen}{RGB}{236,247,240}

\definecolor{mnameblue}{RGB}{243,237,253}

\definecolor{methodpurple}{RGB}{76,40,130}

\newcommand{\dottedlegend}[3][black]{%
  \tikz[baseline=-0.5ex]{
    \draw[#1, line width=#2] (0,0) -- (0.2in,0);
    \fill[#1] (0.1in,0) circle (#3);
  }%
}
\newcolumntype{L}[1]{>{\raggedright\let\newline\\\arraybackslash\hspace{0pt}}m{#1}}
\newcolumntype{C}[1]{>{\centering\let\newline\\\arraybackslash\hspace{0pt}}m{#1}}
\newcolumntype{R}[1]{>{\raggedleft\let\newline\\\arraybackslash\hspace{0pt}}m{#1}}
\newcolumntype{Y}{>{\centering\arraybackslash}X}

\title{\textcolor{methodpurple}{\mname}: Multi-Agent World Models Beyond What Agents See}
\author[1,2]{Runjia Li}
\author[1]{Kam Woh Ng}
\author[2]{Minghao Chen}
\author[1,2]{Junlin Han}
\author[1,2]{Jianyuan Wang}
\author[1]{Zhaochong An}
\author[1]{Juan-Manuel Perez-Rua}
\author[1,2,\dagger]{Andrea Vedaldi}
\author[1,\dagger]{Shikun Liu}

\affiliation[1]{Meta AI}
\affiliation[2]{University of Oxford}

\contribution[\dagger]{Equal supervision.}
\abstract{Despite the success of video world models, extending them to multiple agents is difficult because it requires generating several agent-specific video streams that are consistent in time and between agents.
To address this gap, we introduce \emph{\mname}, a multi-agent world model that augments egocentric agent views with auxiliary camera views.
We suggest that, by fixing the auxiliary views in space, they act as a proxy for a world-centric representation:
Unlike the agents' views, which can change entirely due to a simple change in agent position, changes in the auxiliary views directly map to localized changes in the world state.
We show that these fixed views substantially help the model to maintain consistency.
We also show that, due to their redundant and semi-static nature, the auxiliary views can be efficiently compressed with a $\delta$-VAE that encodes only the changes relative to a static reference image, reducing the number of tokens required to process them.
Experiments in Minecraft and Doom demonstrate improved world and agent consistency, with qualitative results on real-world multi-camera captures indicating that the approach is not tied to game engines.
Our compression analysis further shows that $\delta$-VAE substantially reduces the computational overhead of auxiliary-view generation.
We conclude that these fixed views provide a simple and efficient way to improve multi-agent world models.
}

\date{\today}
\metadata[\raisebox{-0.15em}{\includegraphics[height=1.1em]{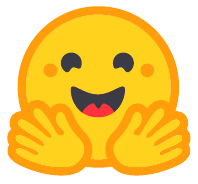}}\hspace{0.35em}Dataset]{\href{https://huggingface.co/collections/liguang0115/overseer}{\nolinkurl{huggingface.co/collections/liguang0115/overseer}}}
\metadata[\raisebox{-0.15em}{\includegraphics[height=1.1em]{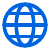}}\hspace{0.35em}Project Page]{\href{https://runjiali-rl.github.io/OverSeer/}{\nolinkurl{overseer-wm.github.io}}}

\begin{document}
\maketitle
\begin{figure*}[h]
\centering
\includegraphics[width=\textwidth]{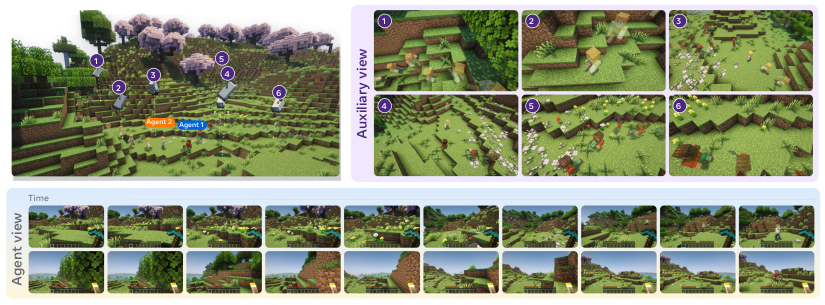}
\caption{\textbf{\mname~sees beyond what agents see.}
Two agents interact in a shared Minecraft environment (Left).
In addition to generating each agent's egocentric view (Bottom), \mname~ models six fixed-pose auxiliary cameras (Right).
These auxiliary streams establish a persistent shared reference, enforcing world consistency over time and agent consistency across views.
}
\label{fig:teaser}
  \vspace{-0.4cm}
\end{figure*}

\section{Introduction}%
\label{sec:intro}

World models aim to learn predictive representations of environments.
Video world models represent the environment as a sequence of images captured from the viewpoint of an acting agent, predicting future frames conditioned on past observations and agent actions.
By taking advantage of large video datasets, these models can learn to represent complex visual dynamics and interactions between characters and objects.
Recent video world models have made substantial progress in visual fidelity and character controllability, but maintaining a coherent world over long horizons remains challenging.
Small prediction errors can accumulate over time, causing objects to drift and environments to change inconsistently.

These challenges become more pronounced in multi-agent environments.
A multi-agent world model must represent not only how the environment evolves but also how multiple characters interact within the same shared world.
We identify two forms of consistency that are essential for this setting.
First, \emph{world consistency} requires persistent scene elements and the effects of past events to be preserved over time: when an agent revisits a location, the scene should reflect its history.
Second, \emph{agent consistency} requires the actions and states of different agents to agree across their observations: an interaction should be depicted consistently in all views from which it is visible.
These constraints are difficult to satisfy using only player-centric streams, because each player observes only a \emph{partial and changing projection of the underlying world}.

One way to address these challenges is to use a \emph{world-centric representation} that explicitly maintains a shared state independent of any individual agent's viewpoint.
Such representations can also model dynamics more efficiently: localized updates occur only when the environment actually changes, whereas an agent's camera motion drastically alters egocentric pixels even in a completely static scene.
However, explicitly reconstructing and maintaining a world-centric representation from partial observations remains challenging.

In this paper, we propose a video-based framework that captures the key advantages of a world-centric representation without requiring explicit global state reconstruction.
To this end, we introduce \emph{\mname}, a multi-agent video world model augmented with \emph{fixed-pose auxiliary camera views} that serve as a practical proxy for a shared world-centric state.
At inference, the auxiliary views are automatically, jointly, and efficiently generated with the agents' views through shared attention.
This design has three benefits: 
\begin{enumerate}[label=(\roman*)]
    \item Auxiliary views cover regions beyond individual agents' immediate fields of view, establishing a {\it shared visual reference} that grounds cross-agent interactions;
    \item Because the auxiliary cameras are stationary, any temporal variation in their views reflects {\it actual environmental changes} rather than camera ego-motion;
    \item  Fixed viewpoints make temporal redundancies predictable and easy to model, reinforcing {\it structural coherence across streams}.
\end{enumerate}

To minimize computational overhead and exploit this inherent temporal redundancy, we introduce $\delta$-VAE, a variational autoencoder that performs \emph{second-stage latent compression} on top of a standard video VAE\@.
For each auxiliary stream, $\delta$-VAE retains an initial reference frame latent and encodes subsequent frames into compact residual latents conditioned on that reference.
The static reference frame preserves persistent background geometry, enabling the residual latents to concentrate exclusively on dynamic events, such as agent movements, object manipulations, and environmental state changes.
By eliminating redundant encoding of unchanged background content, six auxiliary views incur only 37.5\% of the token overhead of the two agent streams.

We evaluate \mname{} on Minecraft and Doom, two games with substantially different visual
dynamics, spatial structures, and interaction patterns, and additionally show qualitative
results on real-world multi-camera captures~\citep{shapovalov23replay}. When agents lose
sight of each other and later reunite, existing multi-agent world models restore the partner
or the scene correctly in only $0$--$12.5\%$ of episodes, whereas \mname{} does so in
$38$--$60\%$, at a small fraction of the cost of modeling the auxiliary views as full video
streams.

\section{Related Work}%
\label{sec:related}

\paragraph{Video Diffusion and Video World Models.}

Diffusion models have emerged as the dominant framework for video generation, evolving from early techniques that augmented pretrained 2D priors with temporal layers~\citep{singer2022makeavideo,blattmann2023videoldm} to large-scale architectures scaled for high-fidelity synthesis~\citep{OpenAI2024Sora,deepmind2025veo3,HaCohen2025LTXVideo,wan2025}.
Video world models extend this paradigm by casting environment simulation as action-conditioned video diffusion, where future frame generation is guided by past observations and agent actions to yield interactive simulators~\citep{bruce2024genie,valevski2024gamengen}.
Recent full-stack systems further enrich this formulation by integrating explicit camera control, scene memory, and real-time streaming~\citep{dreamxteam2026dreamxworld,jiang2026abotworld0,wang2026matrixgame3}.

However, progress in this domain mainly focuses on expanding the input conditioning space, while output generation remains restricted to rendering a single egocentric stream for a single agent.
In contrast, \mname shifts focus to the model's output space: given initial frames and actions from all agents, the model jointly predicts every agent's egocentric stream \emph{alongside} the fixed auxiliary camera views, improving both \textit{world and agent consistency}.

\paragraph{Long-Horizon Consistency and Memory.}

Autoregressive rollouts in video world models tend to compound error and spatial drift over extended horizons~\citep {chen2024diffusionforcing,huang2025selfforcing}.
The prevailing remedy is expanding context capacity through positional extrapolation or streaming attention~\citep{yang2026longlive,yesiltepe2026infinityrope}, memory-bank retrieval by estimated camera pose~\citep{yu2025contextasmemory,hong2025relic,yu2026mosaicmem}, and geometry-guided retrieval~\citep{dreamxteam2026dreamxworld,li25vmem}.
However, these memory structures are constructed entirely from an agent's own past observations. Consequently, unobserved regions are never stored, and retrieval accuracy remains bound to pose estimates that drift alongside generation. 
Our pose-fixed auxiliary views serve a complementary function: while  memory recalls previously observed regions, auxiliary streams provide active coverage of unobserved space.
Furthermore, because these auxiliary views are jointly predicted rather than retrieved, they eliminate reliance on fragile pose estimation altogether.

\paragraph{Multi-Agent and Multi-View World Models.}

Multi-agent world models were first studied in reinforcement learning with compact state spaces~\citep{egorov2022mamba,zhang2025marie}, where multi-view consistency is enforced within an unrendered latent state.
Recent work extends this requirement to the pixel domain, conditioning video diffusion models on multiple action streams to synthesize concurrent agent views ~\citep{enigma2025multiverse,savva2026solaris,liu2026gammaworld,mira2026,wu2026multiworld,hu2026metaworld}. Multi-view generation also appears in generative 3D modeling and autonomous driving~\citep{kong2024eschernet,liu2026kaleido,jang2026raxel,gao2025cat3d,gao2024magicdrive,gao2024vista}, though cameras in these settings are typically in a static scene or camera rigs rigidly attached to a single moving platform.

Existing approaches to multi-agent consistency broadly fall into two categories. 
The first couples egocentric streams via stream concatenation, shared attention, tiling, or permutation-symmetric encodings~\citep{enigma2025multiverse,savva2026solaris,mira2026,liu2026gammaworld}, meaning every reference frame is itself a moving, partial observation. 
The second improves consistency via an external state, such as an editable map with estimated per-agent poses~\citep{po2026multigen}; while this decouples the reference from individual agent trajectories, it models spatial geometry rather than visual appearance and remains vulnerable to pose drift. 

In contrast, \mname improves consistency by introducing auxiliary streams that are jointly \textit{predicted} alongside the agents' streams rather than supplied as external map inputs. 

\paragraph{Efficient Video Tokenization.}

Modern video generators typically operate within the latent space of a video VAE~\citep{wan2025,HaCohen2025LTXVideo}, with recent efficiency efforts compressing representations further via stronger tokenizers~\citep{yu2024magvitv2,chen2025dcae}, token merging~\citep{bolya2022tokenmerging}, lightweight decoders, and bounded KV caches~\citep{jiang2026abotworld0,mira2026}.
However, these generic techniques allocate computational capacity uniformly across all streams, causing an auxiliary camera view to incur nearly the same computational cost as an egocentric agent stream. 

A closer line of work changes what a latent space \emph{represents}: neural video codecs encode frames relative to a reference frame via residuals~\citep{lu2019dvc} or by conditioning~\citep{li2021dcvc}, RefTok tokenizes each frame relative to a reference frame~\citep{fan2025reftok}, and DeltaTok compresses consecutive-frame feature differences into a single delta token~\citep{kerssies2026deltatok}.
In these approaches, the reference frame is not fixed, but updated regularly over time, missing the advantages of \mname's fixed views, starting from the fact that they can be compressed effectively.

\section{\mname}%
\label{sec:method}

\subsection{Background and Notation}

\paragraph{Video World Modeling.}

Video world models usually represent the evolution of the world from the viewpoint of an agent.
Let $o_t$ denote the visual observation at time $t$ (i.e., an image), $a_t$ the agent action, and $z_t=\mathcal{E}(o_t)$ a learned latent representation of the observation, where $\mathcal{E}$ denotes a visual encoder.
Let $s_t$ denote the world state available to the model at time $t$.
A general one-step video world model can be formulated as:
\begin{align}
    \label{eq:world_model}
    p(z_{t+1}\mid s_t,a_t)
\end{align}
Depending on the model, $s_t$ may correspond to the current representation $z_t$, a history of representations $z_{\leq t}$, or a learned summary of the observation-action history $f(z_{\leq t},a_{<t})$.

\paragraph{Multi-Agent Video World Modeling.}
Consider an environment with $N$ agents. We use the superscript $A$ to denote agent-centric
camera streams. Agent $i$ receives an observation $o_t^{A,i}$, represented as
$z_t^{A,i}=\mathcal{E}(o_t^{A,i})$, and executes an action $a_t^i$. We denote the joint
agent observations, representations, and actions by $\mathbf{o}_t^A=\{o_t^{A,i}\}_{i=1}^{N}$,
$\mathbf{z}_t^A=\{z_t^{A,i}\}_{i=1}^{N}$, and $\mathbf{a}_t=\{a_t^i\}_{i=1}^{N}$. A
multi-agent video world model extends \cref{eq:world_model} to joint prediction:
\begin{align}
    \label{eq:ma_world_model}
    p(\mathbf{z}^A_{t+1}\mid \mathbf{s}_t,\mathbf{a}_t),
\end{align}
where $\mathbf{s}_t$ represents the shared world state. Factorizing this into
independent single-agent models, $\prod_i p(z^{A,i}_{t+1}\mid s^i_t,a^i_t)$, is
insufficient: each $s^i_t$ is built only from agent $i$'s own history, so nothing enforces
that agents agree where their views overlap, or that a region seen by one agent stays
consistent when another later observes it. The challenge is that $\mathbf{s}_t$ must be
inferred from $\mathbf{z}^A_{\le t}$, which are partial and change as the agents move.

\paragraph{Auxiliary Camera Views.}

To provide persistent spatial references for the shared environment, \mname introduces $K$ fixed-pose auxiliary camera views, denoted by $\mathbf{o}_t^X=\{o_t^{X,j}\}_{j=1}^{K}$, with corresponding camera poses $\mathbf{c}=\{c^j\}_{j=1}^{K}$, where the superscript $X$ denotes auxiliary camera streams.
Unlike agent-centric cameras, whose viewpoints change with agent motion, the auxiliary cameras remain fixed over time.
These views change proportionally to changes in the underlying world's state and provide a stable spatial reference for all the agents.
During training, the auxiliary views also teach the model what happens outside the agents' immediate purview.

\subsection{\mname~Model Design}%
\label{sec:model_design}

\begin{figure*}[t]
  \centering
  \includegraphics[width=\textwidth]{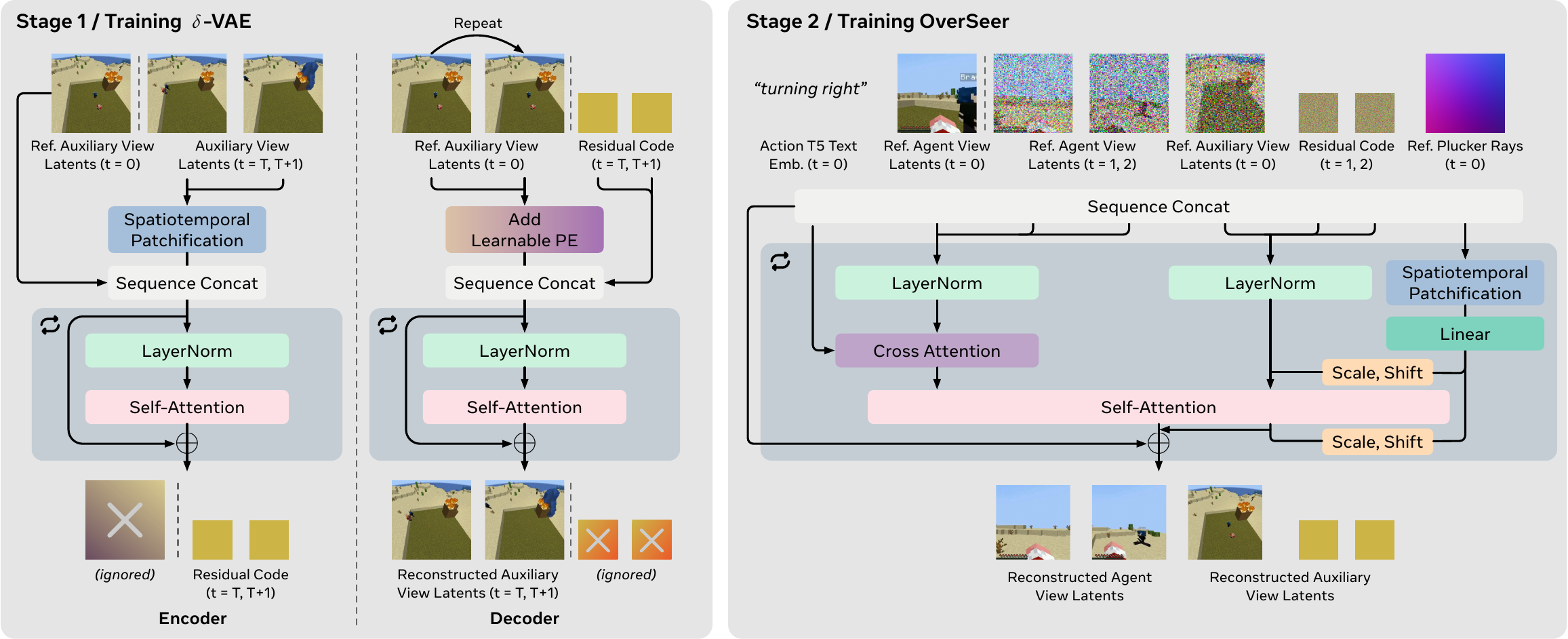}
\caption{\textbf{\mname~design overview.} A $\delta$-VAE (Left) compresses each fixed auxiliary stream into compact residual codes relative to its first-frame reference, and OverSeer (Right) jointly generates agent latents and these residual codes, conditioned on T5-encoded actions via cross-attention and Plücker-ray camera poses via AdaLN.}
  \label{fig:method}
\end{figure*}

We design \mname as a latent video diffusion model~\citep{rombach2022ldm} built upon Wan2.2-TI2V-5B~\citep{wan2025}, leveraging its strong pre-trained video generation prior.
An overview is shown in \cref{fig:method}.
Following the standard latent diffusion design, input visual observations are mapped into a compact latent space using a pre-trained video VAE~\citep{wan2025}, yielding agent latents $\mathbf{z}_t^A=\{\mathcal{E}(o); \forall o\in\mathbf{o}_t^A\}$ and auxiliary latents $\mathbf{z}_t^X=\{\mathcal{E}(o);\forall o\in\mathbf{o}_t^X\}$.

To keep the latent sequence lengths tractable during diffusion, we do not model auxiliary latents frame-by-frame.
Instead, each auxiliary stream is represented by its first-frame latent as a fixed anchor, $\mathbf{z}_0^X=\{z_0^{X,j}\}_{j=1}^{K}$, paired with compact residual codes $\mathbf{d}^X_{1:Q}=\{d^{X,j}_{1:Q}\}_{j=1}^K$, a length-$Q$ sequence per view that encodes subsequent temporal dynamics produced by $\delta$-VAE (see \cref{sec:delta_vae}).
Extending the world-model formulation in \cref{eq:world_model} to multi-agent generation with auxiliary views, \mname jointly models the future agent latents, initial auxiliary anchor frames, and compressed auxiliary dynamics, conditioned on the initial agent frames and actions, and auxiliary camera poses:
\begin{align}
p\left(
\mathbf{z}_{1:T}^A,
\mathbf{z}_0^X,
\mathbf{d}_{1:T}^X
\mid
\mathbf{z}_0^A,
\mathbf{a}_{0:T-1},
\mathbf{c}
\right).
\label{eq:joint_world_model}
\end{align}

\paragraph{Training and Inference.}

Knowledge of the ground-truth auxiliary visual streams is only required during training. At inference, \mname{} receives the initial agent observations and agent actions along with auxiliary camera poses. The auxiliary camera poses are sampled from a \emph{scene-independent prior} estimated from training poses and greedily selected to maximize field-of-view coverage without scene-specific input (see \cref{app:pose_sampling}).
The model jointly generates future agent observations and the auxiliary streams, including their initial anchor frames.
Ground-truth auxiliary images are used only during training.

We train \mname using diffusion forcing~\citep{chen2024diffusionforcing}, where diffusion timesteps are sampled independently across temporal positions, and target latents are noised accordingly while initial conditioning frames $\mathbf{z}_0^A$ remain clean.
All latents, including agent latents, auxiliary anchor frames, and residual codes, are projected to the same DiT hidden dimension with learnable view-identification embeddings added to each view's tokens for differentiation.

Agent and auxiliary streams are governed by different modalities and therefore use complementary conditioning pathways.
For each agent stream, discrete actions (including agent navigation, camera motions, and inventory interactions) are formatted into short textual descriptions, encoded via a frozen T5 text encoder~\citep{raffel2020t5}, and injected into the DiT layers through cross-attention.
For each auxiliary stream, camera pose $c^j$ and its known intrinsics are converted into a spatial Pl\"ucker-ray map~\citep{sitzmann2021lfns}.
These ray representations are patchified and projected into Adaptive Layer Normalization (AdaLN)~\citep{perez18film:}, with scale and shift parameters to modulate the corresponding auxiliary tokens.

Tokens from all agents and auxiliary streams are then jointly processed by shared self-attention layers, enabling direct cross-modality communication.
We enforce a causal temporal attention mask throughout the network: tokens at a given temporal position $t$ can attend only to conditioning tokens and to tokens from identical or earlier temporal positions ($t' \le t$).
This prevents future latent states from leaking into the prediction of earlier states under independent diffusion forcing timesteps.

\subsection{Efficient Modeling of Auxiliary Dynamics}%
\label{sec:delta_vae}
\begin{figure*}[t]
\centering
\includegraphics[width=\textwidth]{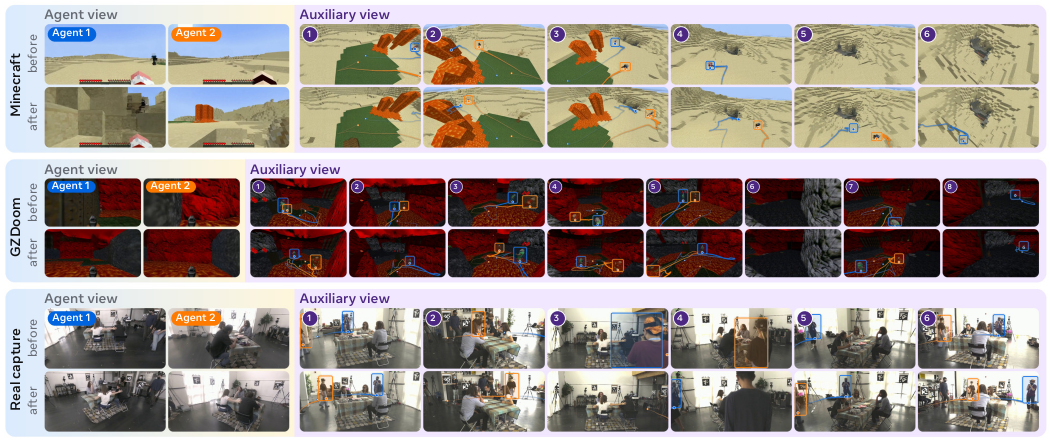}
\caption{\textbf{\mname~keeps the agents consistent across views.} Generated videos in \textbf{Minecraft},
\textbf{GZDoom} and a multi-camera \textbf{Real Capture}, each at two moments (before, after). Each agent's trail is one
3D track projected into every auxiliary camera (Agent~1 blue, Agent~2 orange); it lands on the
generated agent in each of them, so the agents are consistent across views.}%
\label{fig:multiview_consistency}
\end{figure*}

Because the auxiliary cameras are pose-fixed, their background visual content remains predominantly static across time, with scene updates concentrated in localized regions affected by agent activity and environmental dynamics. Directly modeling full latent representations for every auxiliary frame therefore introduces severe temporal and spatial redundancy. 
To address this inefficiency, we introduce $\delta$-VAE, a second-stage variational autoencoder that exploits this background stationarity to further compress the latent streams produced by the primary video VAE\@.

\paragraph{Anchor and Residual Codes.}
For each auxiliary view $j$, we retain the initial video-VAE latent $z_0^{X,j} \in \mathbb{R}^{H \times W \times C}$ as a static reference anchor and partition the subsequent temporal sequence into $Q$ non-overlapping chunks. 
Let $\mathcal{T}_q$ denote the time indices of the target latents in chunk $q$, where $|\mathcal{T}_q| = \tau$ represents the temporal compression factor. 
For each chunk $q$, $\delta$-VAE encodes the anchor $z_0^{X,j}$ alongside the corresponding target latents $\{z_t^{X,j}\}_{t \in \mathcal{T}_q}$ into a compact residual code $d_q^{X,j} \in \mathbb{R}^{\frac{H}{s} \times \frac{W}{s} \times C_d}$, where $s$ denotes the spatial compression factor. 
These residual codes achieve an $s^2 \times$ spatial and $\tau \times$ temporal token reduction relative to the primary video-VAE latents. 
The resulting compressed sequence for auxiliary view $j$ is represented as $\widetilde{z}^{X,j} = [z_0^{X,j}; d_1^{X,j}; \cdots; d_Q^{X,j}]$.

\paragraph{$\delta$-VAE Architecture Design.}

The encoder $\mathcal{E}_{\delta\text{-}\mathrm{VAE}}$ spatially patchifies the video-VAE features and jointly processes the anchor latent $z_0^{X,j}$ and the target latents $\{z_t^{X,j}\}_{t\in\mathcal{T}_q}$ using a spatio-temporal transformer.
It parameterizes a Gaussian latent distribution $\mathcal{G}_q^{X,j}$~\citep{kingma2014vae} from which the residual code $d_q^{X,j}$ is sampled.
The decoder $\mathcal{D}_{\delta\text{-}\mathrm{VAE}}$ reconstructs each target chunk from the residual code $d_q^{X,j}$ and the anchor $z_0^{X,j}$.
The residual code remains at its compressed spatial resolution, while the anchor is repeated over the $\tau$ target temporal positions by adding a learnable temporal positional embedding.
These replicated anchor features then attend to $d_q^{X,j}$ through cross-attention layers, producing the reconstructed target latents $\{\hat{z}_t^{X,j}\}_{t\in\mathcal{T}_q}$.

\paragraph{Training Objective.}

We train the $\delta$-VAE with three losses: an $\ell_1$ reconstruction loss $\mathcal{L}_{\mathrm{rec}}$ between the reconstructed and target video-VAE latents; the standard VAE KL divergence $\mathcal{L}_{\mathrm{KL}}$; and a feature alignment loss $\mathcal{L}_{\mathrm{align}}$ inspired by REPA~\citep{repa}.
Reconstruction alone lets the residual code spend its capacity on static background that the anchor already provides; $\mathcal{L}_{\mathrm{align}}$ instead aligns the code with video-VAE features re-weighted toward dynamic regions.
To build this alignment target, we first compute a spatial change weight map for each target frame from its channel-wise latent distance to the anchor,
{
\small
\begin{align}
\widetilde{M}_t^{X,j}
=
\left\|
z_t^{X,j}-z_0^{X,j}
\right\|_{2,C},
\qquad
M_t^{X,j}
=
\frac{\widetilde{M}_t^{X,j}}{\left\|\widetilde{M}_t^{X,j}\right\|_2},
\end{align}
}%
where $\|\cdot\|_{2,C}$ denotes the $\ell_2$ norm over channels and $M_t^{X,j}\in\mathbb{R}^{H\times W}$ assigns higher weights to spatial locations exhibiting greater temporal variation relative to the anchor frame.
We modulate the target video-VAE features using this weight map via $r_t^{X,j}=M_t^{X,j}\odot z_t^{X,j}$.
Importantly, the latent distance is used solely to isolate dynamic regions; the feature alignment target is constructed directly from the target VAE features themselves, as naive subtraction within a non-linear VAE latent space does not generally yield semantically meaningful representations.

For each chunk, the weighted feature $\{r_t^{X,j}\}_{t\in\mathcal{T}_q}\in\mathbb{R}^{\tau\times H\times W\times C}$ is pooled over $\tau\times s\times s$ temporal-spatial blocks, yielding a target feature representation in $\mathbb{R}^{\frac{H}{s}\times\frac{W}{s}\times C}$.
A learned linear projection $\mathcal{P}$ maps $d_q^{X,j}$ from $C_d$ to $C$, establishing the feature alignment loss. Combining this with the $\ell_1$ reconstruction loss $\mathcal{L}_{\mathrm{rec}}$ and the KL divergence regularization $\mathcal{L}_{\mathrm{KL}}$, the loss terms are defined as:
{
\begin{align}
\mathcal{L}_{\mathrm{rec}} = \|\hat{z}_t^{X,j} - z_t^{X,j}\|_1, \
\mathcal{L}_{\mathrm{KL}} = \mathrm{KL}\big(\mathcal{G}_q^{X,j} \,\|\, \mathcal{N}(\mathbf{0}, \mathbf{I})\big), \
\mathcal{L}_{\mathrm{align}}
=
\left \|
\mathcal{P}\left(d_q^{X,j}\right)
-
\operatorname{Pool}
\left(
\{r_t^{X,j}\}_{t\in\mathcal{T}_q}
\right)
\right \|_2^2.
\end{align}
}%
The overall training objective for $\delta$-VAE is: $
\mathcal{L}_{\delta\text{-}\mathrm{VAE}}
=
\lambda_{\mathrm{rec}}\mathcal{L}_{\mathrm{rec}}
+
\lambda_{\mathrm{KL}}\mathcal{L}_{\mathrm{KL}}
+
\lambda_{\mathrm{align}}\mathcal{L}_{\mathrm{align}}
$
with loss weights specified in \cref{app:impl}.
We first train the $\delta$-VAE individually on auxiliary-camera streams and freeze its weights prior to training the primary world model \mname.
During \mname training, ground-truth auxiliary videos are encoded into anchor latents $\mathbf{z}_0^X$ and residual codes $\mathbf{d}_{1:Q}^X$, which serve as diffusion targets alongside future agent latents $\mathbf{z}_{1:T}^A$ in \cref{eq:joint_world_model}.
\begin{figure*}[t]
  \centering
  \includegraphics[width=\textwidth]{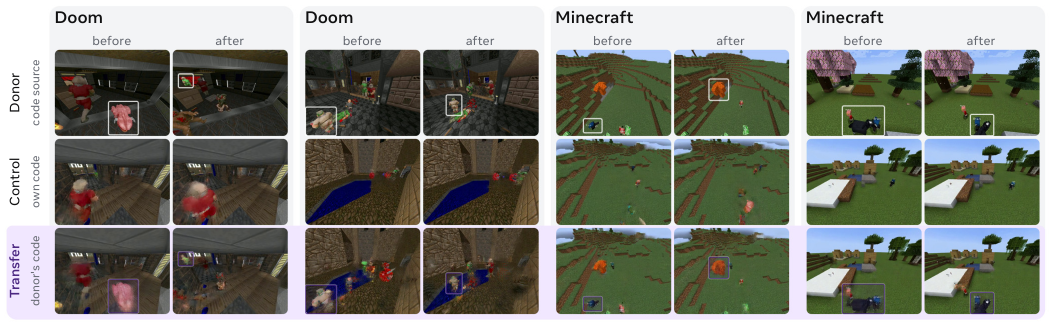}
\caption{\textbf{$\delta$-VAE residual codes carry the movers, not the scene.} Four cross-scene
  transfers: a host anchor is decoded with a donor clip's residual code (transfer, purple band) instead of
  its own (control), at two moments (before, after). The donor's mover (white box) appears in the host
  scene (purple box) while the host scene itself is preserved.}
  \label{fig:transfer_examples}
\end{figure*}

\section{Experiments}%
\label{sec:experiments}

We structure our experimental evaluation to test the central hypothesis of this work: extending a multi-agent video world model with pose-fixed, efficiently represented auxiliary views significantly improves both temporal and agent consistency compared to baselines (\cref{sec:exp_main}).
We then conduct ablation studies investigating how performance scales with the number of auxiliary views and how the $\delta$-VAE compression ratio influences synthesis quality and computational overhead  (\cref{sec:exp_ablation}).
For qualitative results, please refer to our website and appendix.

\paragraph{Environments and Benchmarks.}
We evaluate world models by conditioning future frame rollouts on initial visual observations and multi-agent action trajectories. Primary quantitative evaluations are conducted in \textbf{Minecraft}, whose persistent and editable 3D world naturally tests spatial memory, multi-agent interaction, and cross-view consistency (\cref{tab:minecraft_main,tab:aux_delta_ablation}). To evaluate long-horizon spatial reasoning, we additionally run {\it revisit} tasks in \textbf{Doom} (\cref{tab:aux_delta_ablation}) and showcase qualitative generalization on real-world video captures from \textbf{Replay}~\citep{shapovalov23replay}. In Minecraft, evaluation spans two benchmark suites:
\begin{enumerate}[leftmargin=*]
    \item \textbf{Solaris Benchmark}~\citep{savva2026solaris}: Evaluates five multi-agent capabilities using VLM-judged scripted questions, requiring all questions within an episode to be answered correctly.
    \item \textbf{\benchname{} (Multi-Agent Consistency Benchmark)}: Designed specifically to evaluate world and agent consistency in a harder setting. Specifically. we design the following three tasks:
    \begin{itemize}[leftmargin=*,topsep=0pt]
        \item \textit{Reunite at Start}: Both agents return to their exact initial poses. Consequently, an agent's partner must reappear at a previously observed spatial location.
        \item \textit{Reunite Far}: Agents navigate long, converging routes without visual contact in transit, turning to face each other upon arrival. A partner must appear at close range within previously unobserved space.
        \item \textit{Dynamic Event}: Agents observe the initial stage of a localized dynamic process ({\it e.g.}, a chair catching fire) before separating. Upon return, the object state must reflect correct off-screen temporal evolution.
    \end{itemize}
\end{enumerate}
In Doom, we design two \textit{Reunite} tasks across six distinct maps with extended navigation paths.

\paragraph{Evaluation Metrics.}
We report VLM Episode Accuracy (VLM\textuparrow) and Frame Fr\'echet Inception Distance (FID\textdownarrow). VLM is evaluated using Claude Opus 5 conditioned on ground-truth camera poses and environment event logs. Metrics are averaged across $32$ episodes per task in Minecraft and $100$ episodes for each Doom task.

\paragraph{Baselines.}
We compare \mname{} against two state-of-the-art two-player world models for Minecraft: Solaris~\citep{savva2026solaris} and $\gamma$-World~\citep{liu2026gammaworld}. For a fair comparison, all models are evaluated with their non-distilled, causally fine-tuned checkpoints and receive identical conditioning frames and action sequences, operating strictly without auxiliary views at inference.

\begin{table*}[t]
\centering
\footnotesize
\setlength{\tabcolsep}{3.3pt}
\renewcommand{\arraystretch}{1.08}

\begin{tabular}{@{}l*{16}{c}@{}}
\toprule

& \multicolumn{10}{c}{\textbf{Solaris Benchmark}}
& \multicolumn{6}{c}{\textbf{\benchname}} \\
\cmidrule(lr){2-11}
\cmidrule(lr){12-17}

& \multicolumn{2}{c}{\textbf{Movement}}
& \multicolumn{2}{c}{\textbf{Grounding}}
& \multicolumn{2}{c}{\textbf{Memory}}
& \multicolumn{2}{c}{\textbf{Building}}
& \multicolumn{2}{c}{\textbf{Consistency}}
& \multicolumn{2}{c}{\textbf{Reunite at Start}}
& \multicolumn{2}{c}{\textbf{Reunite Far}}
& \multicolumn{2}{c}{\textbf{Dynamic Event}} \\
\cmidrule(lr){2-3}
\cmidrule(lr){4-5}
\cmidrule(lr){6-7}
\cmidrule(lr){8-9}
\cmidrule(lr){10-11}
\cmidrule(lr){12-13}
\cmidrule(lr){14-15}
\cmidrule(lr){16-17}

\textbf{Method}
& \textbf{VLM$\uparrow$} & \textbf{FID$\downarrow$}
& \textbf{VLM$\uparrow$} & \textbf{FID$\downarrow$}
& \textbf{VLM$\uparrow$} & \textbf{FID$\downarrow$}
& \textbf{VLM$\uparrow$} & \textbf{FID$\downarrow$}
& \textbf{VLM$\uparrow$} & \textbf{FID$\downarrow$}
& \textbf{VLM$\uparrow$} & \textbf{FID$\downarrow$}
& \textbf{VLM$\uparrow$} & \textbf{FID$\downarrow$}
& \textbf{VLM$\uparrow$} & \textbf{FID$\downarrow$} \\
\midrule

\textbf{Solaris}
& \textbf{68.8} & 22.38
& \underline{66.7} & \underline{14.92}
& \underline{20.8} & 47.11
& 22.9 & 76.01
& \underline{56.3} & 68.42
& \underline{0.0} & \underline{17.06}
& \underline{6.2} & \underline{20.23}
& \underline{12.5} & \underline{26.52} \\

\textbf{$\gamma$-World}
& \underline{64.6} & \textbf{17.13}
& 62.5 & 29.39
& 16.7 & \underline{29.57}
& \textbf{43.8} & \textbf{24.62}
& 47.9 & \underline{56.39}
& \underline{0.0} & 23.83
& 0.0 & 25.71
& 8.3 & 31.72 \\

\rowcolor{mnameblue}
\textbf{\mname}
& 59.1 & \underline{18.27}
& \textbf{91.7} & \textbf{11.33}
& \textbf{87.5} & \textbf{12.95}
& \underline{41.7} & \underline{32.50}
& \textbf{60.4} & \textbf{26.47}
& \textbf{60.2} & \textbf{13.99}
& \textbf{37.9} & \textbf{12.79}
& \textbf{40.5} & \textbf{17.61} \\

\bottomrule
\end{tabular}

\caption{\textbf{Results on Minecraft.}
Comparison on the Solaris benchmark and our \benchname{}. VLM denotes VLM-judged accuracy (\%) and FID denotes frame-level FID; \benchname{} results are averaged over 5 runs.
\textbf{Bold} and \underline{underline} denote the best and second-best
results in each column, respectively.}
\label{tab:minecraft_main}
\end{table*}

\subsection{Analysis of Multi-Agent World Modeling Benchmarks}%
\label{sec:exp_main}

\paragraph{Main results.}

\begin{figure*}[t]
  \centering
  \includegraphics[width=\textwidth]{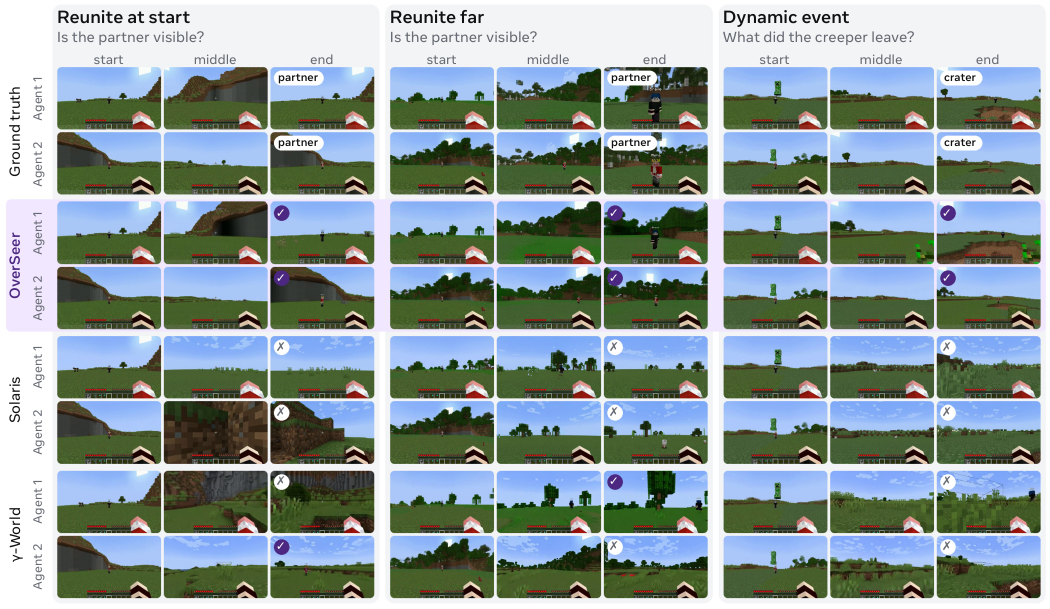}
  \caption{\textbf{\mname~keeps the world consistent over time.} On the three revisiting tasks
  (reunite at start, reunite far, dynamic event), \mname~shows the partner and the crater where the
  ground truth does, while Solaris and $\gamma$-World mostly do not.}
  \label{fig:campaign_tasks}
\end{figure*}

As shown in \cref{tab:minecraft_main}, \mname{} leads on grounding, memory, and consistency, the Solaris capabilities that require spatial reasoning beyond the current view. Conversely, it trails the best baseline on building and movement tasks, which depend primarily on short-horizon egocentric streams. On \benchname{}, \mname{} yields substantial gains in VLM accuracy over prior baselines, increasing scores from $0.0\% / 6.2\% / 12.5\%$ to $60.2\% / 37.9\% / 40.5\%$ across our three designed tasks (also qualitatively shown in \cref{fig:campaign_tasks}), while achieving the lowest (best) FID score across all tasks. Agent consistency is further demonstrated qualitatively in \cref{fig:multiview_consistency}: projected 3D agent trajectories precisely align with generated agents across every auxiliary view in Minecraft, Doom, and real-world captures. Extended visualization rollouts are provided in \cref{appendix:more_vis}.

\paragraph{FID does not reflect generation consistency.}

On \emph{Reunite at Start}, Solaris obtains an FID of $17.06$, within $3.1$ points of \mname, yet answers none of the episodes correctly.
While its generated rollouts remain {\it visually in-distribution}, they fail to reconstruct the specific scene previously left behind by the agents, frequently filling in existing craters or omitting partner agents entirely (\cref{fig:campaign_tasks}). 
This disconnect demonstrates that standard frame-level distributional metrics like FID are insensitive to structural memory failures, motivating the content-level queries proposed in our new benchmark.

\paragraph{Without a shared view, agents drift apart.}

Both baselines enforce cross-agent agreement by coupling the agent streams directly, through concatenation or shared attention.
Under this design, every reference frame is itself a dynamic, partial observation, restricting agreement exclusively to moments when the agents' fields of view overlap. 
Empirical results directly reflect this structural limitation: once agents lose visual contact, $\gamma$-World does not achieve any of the reunite tasks. 
The same argument accounts for the gap between \emph{Reunite at Start} ($60.2$) and \emph{Reunite Far} ($37.9$), as sampled cameras cover the start region more densely than the periphery.

\subsection{Ablations}%
\label{sec:exp_ablation}

\paragraph{Effect of $\delta$-VAE.}
Removing $\delta$-VAE improves visual quality and, on Doom, consistency as well
(\cref{tab:aux_delta_ablation}), but at the cost of an $8\times$ larger auxiliary-stream token footprint  ($864$ vs.\ $108$ tokens per frame at $576\times384$). This performance trade-off is domain-dependent: a residual code remains faithful only insofar as the scene stays close to its static anchor. Doom's rapid lighting changes and enemy movements generate much larger residual updates than Minecraft's slow, blocky dynamics.

\begin{table*}[ht!]
\centering
\footnotesize
\setlength{\tabcolsep}{4.2pt}
\renewcommand{\arraystretch}{1.05}

\begin{tabular}{@{}l*{10}{c}@{}}
\toprule
& \multicolumn{6}{c}{\textbf{Minecraft}}
& \multicolumn{4}{c}{\textbf{Doom}} \\
\cmidrule(lr){2-7}
\cmidrule(lr){8-11}

& \multicolumn{2}{c}{\textbf{Reunite at Start}}
& \multicolumn{2}{c}{\textbf{Reunite Far}}
& \multicolumn{2}{c}{\textbf{Dynamic Event}}
& \multicolumn{2}{c}{\textbf{Reunite at Start}}
& \multicolumn{2}{c}{\textbf{Reunite Far}} \\
\cmidrule(lr){2-3}
\cmidrule(lr){4-5}
\cmidrule(lr){6-7}
\cmidrule(lr){8-9}
\cmidrule(lr){10-11}

\textbf{Variant}
& \textbf{VLM$\uparrow$} & \textbf{FID$\downarrow$}
& \textbf{VLM$\uparrow$} & \textbf{FID$\downarrow$}
& \textbf{VLM$\uparrow$} & \textbf{FID$\downarrow$}
& \textbf{VLM$\uparrow$} & \textbf{FID$\downarrow$}
& \textbf{VLM$\uparrow$} & \textbf{FID$\downarrow$} \\
\midrule

\rowcolor{mnameblue}
\textbf{\mname}
& \textbf{60.2} & 14.0
& 37.9 & \textbf{12.8}
& 40.5 & 17.6
& 62.0 & 23.2
& 67.0 & 20.9 \\
\midrule
w/o Aux. Views
& 25.0 & 24.6
& 16.7 & 27.2
& 20.8 & 33.1
& 12.0 & 50.6
& 0.0 & 87.4 \\

w/o $\delta$-VAE
& 59.4 & \textbf{8.2}
& \textbf{40.6} & 13.4
& \textbf{41.7} & \textbf{10.3}
& \textbf{78.0} & \textbf{15.8}
& \textbf{84.0} & \textbf{15.4} \\

\bottomrule
\end{tabular}%

\caption{\textbf{Ablation of auxiliary views and $\delta$-VAE.}
Comparison on \benchname{} in Minecraft and Doom. w/o $\delta$-VAE models auxiliary views with uncompressed video-VAE latents, using $8\times$ more auxiliary tokens.
\textbf{Bold} denotes the best result in each column.}
\label{tab:aux_delta_ablation}
\end{table*}

\begin{figure}[ht!]
  \centering
    \begin{subfigure}[b]{0.48\linewidth}
      \centering
       {\footnotesize \dottedlegend[Maroon]{1pt}{2pt} Movement \quad \dottedlegend[Peach]{1pt}{2pt} Grounding \quad \dottedlegend[Fuchsia]{1pt}{2pt} Memory \\
       \dottedlegend[NavyBlue]{1pt}{2pt} Building \quad \dottedlegend[ForestGreen]{1pt}{2pt} Consistency } 
        \resizebox{0.49\linewidth}{!}{{\begin{tikzpicture}

\definecolor{c1}{RGB}{214,39,40}
\definecolor{c2}{RGB}{255,127,14}
\definecolor{c3}{RGB}{44,160,44}
\definecolor{c4}{RGB}{148,103,189}
\definecolor{c5}{RGB}{31,119,180}

\begin{axis}[
    every axis y label/.style={at={(current axis.north west)},above=0mm},
    legend cell align={left},
    legend style={fill opacity=0.0, draw opacity=1, text opacity=1, draw=none, legend columns=1, column sep=2pt, very thick, anchor=north west, at={(0.0,1.0)}, font=\footnotesize},
    width=6cm, height=4cm,
    tick align=outside,
    tick pos=left,
    xmin=0, xmax=6.2,
    ymin=5, ymax=80,
    xlabel={\# Auxiliary Cameras},
    xtick={0, 1, 2, 4, 6},
    ytick={10, 30, 50, 70},
    ylabel={FID \textdownarrow},
]
\addplot [thick, mark=*, Maroon]
table {%
0 16.6
1 18.51
2 17.93
4 16.65
6 18.27
};
\addplot [thick, mark=*, Peach]
table {%
0 42.67
1 32
2 22.51
4 14.78
6 11.33
};
\addplot [thick, mark=*, Fuchsia]
table {%
0 38.91
1 23.52
2 19.68
4 13.78
6 12.95
};
\addplot [thick, mark=*, NavyBlue]
table {%
0 29.5
1 33.3
2 30.83
4 27.89
6 32.5
};
\addplot [thick, mark=*, ForestGreen]
table {%
0 67.43
1 48.38
2 40.36
4 34.2
6 26.47
};
\draw[gray!65, thin, dashed](axis cs:0,10) -- (axis cs:7,10);
\draw[gray!65, thin, dashed](axis cs:0,30) -- (axis cs:7,30);
\draw[gray!65, thin, dashed](axis cs:0,50) -- (axis cs:7,50);
\draw[gray!65, thin, dashed](axis cs:0,70) -- (axis cs:7,70);
\end{axis}

\end{tikzpicture}}}%
        \resizebox{0.49\linewidth}{!}{{\begin{tikzpicture}

\begin{axis}[
    every axis y label/.style={at={(current axis.north west)},above=0mm},
    legend cell align={left},
    legend style={fill opacity=0.0, draw opacity=1, text opacity=1, draw=none, legend columns=1, column sep=2pt, very thick, anchor=north west, at={(0.0,1.0)}, font=\footnotesize},
    width=6cm, height=4cm,
    tick align=outside,
    tick pos=left,
    xmin=0, xmax=6.2,
    ymin=0, ymax=100,
    xlabel={\# Auxiliary Cameras},
    xtick={0, 1, 2, 4, 6},
    ytick={10, 30, 50, 70, 90},
    ylabel={VLM \textuparrow},
]
\addplot [thick, mark=*, Maroon]
table {%
0 60.42
1 58.33
2 60.42
4 56.25
6 59.1
};
\addplot [thick, mark=*, Peach]
table {%
0 60.42
1 68.75
2 75
4 87.5
6 91.6
};
\addplot [thick, mark=*, Fuchsia]
table {%
0 14.58
1 35.42
2 47.92
4 75
6 87.5
};
\addplot [thick, mark=*, NavyBlue]
table {%
0 37.5
1 39.58
2 37.5
4 39.58
6 41.6
};
\addplot [thick, mark=*, ForestGreen]
table {%
0 45.83
1 47.92
2 50
4 54.17
6 60.4
};
\draw[gray!65, thin, dashed](axis cs:0,10) -- (axis cs:7,10);
\draw[gray!65, thin, dashed](axis cs:0,30) -- (axis cs:7,30);
\draw[gray!65, thin, dashed](axis cs:0,50) -- (axis cs:7,50);
\draw[gray!65, thin, dashed](axis cs:0,70) -- (axis cs:7,70);
\draw[gray!65, thin, dashed](axis cs:0,90) -- (axis cs:7,90);
\end{axis}

\end{tikzpicture}}}
      \caption{\centering Solaris Benchmark}
      \label{fig:cctv_cam}
    \end{subfigure} \hfill \vrule \hfill
   \begin{subfigure}[b]{0.48\linewidth}
      \centering
      {\footnotesize \dottedlegend[Maroon]{1pt}{2pt} Reunite at Start \quad \dottedlegend[ForestGreen]{1pt}{2pt} Reunite Far \\ \dottedlegend[NavyBlue]{1pt}{2pt} Dynamic Event }
      \resizebox{0.49\linewidth}{!}{{\begin{tikzpicture}

\definecolor{darkgray176}{RGB}{176,176,176}
\definecolor{darkorange25512714}{RGB}{255,127,14}
\definecolor{forestgreen4416044}{RGB}{44,160,44}
\definecolor{steelblue31119180}{RGB}{31,119,180}

\begin{axis}[
    every axis y label/.style={at={(current axis.north west)},above=0mm},
    legend cell align={left},
    legend style={fill opacity=0.0, draw opacity=1, text opacity=1, draw=none, legend columns=1, column sep=2pt, very thick, anchor=north west, at={(0.0,1.0)}, font=\footnotesize},
    width=6cm, height=4cm,
    tick align=outside,
    tick pos=left,
    xmin=0, xmax=6.2,
    ymin=5, ymax=45,
    xlabel={\# Auxiliary Cameras},
    xtick={0, 1, 2, 4, 6},
    ytick={10, 20, 30, 40},
    ylabel={FID \textdownarrow},
]
\addplot [thick, mark=*, Maroon]
table {%
0 24.6
1 18.33
2 16.6
4 14.97
6 13.99
};
\addplot [thick, mark=*, ForestGreen]
table {%
0 27.2
1 22.37
2 19.68
4 16.24
6 12.79
};
\addplot [thick, mark=*, NavyBlue]
table {%
0 33.1
1 26.23
2 22.74
4 19.85
6 17.61
};
\draw[gray!65, thin, dashed](axis cs:0,10) -- (axis cs:7,10);
\draw[gray!65, thin, dashed](axis cs:0,20) -- (axis cs:7,20);
\draw[gray!65, thin, dashed](axis cs:0,30) -- (axis cs:7,30);
\draw[gray!65, thin, dashed](axis cs:0,40) -- (axis cs:7,40);
\end{axis}

\end{tikzpicture}}}%
       \resizebox{0.49\linewidth}{!}{{\begin{tikzpicture}

\definecolor{darkgray176}{RGB}{176,176,176}
\definecolor{darkorange25512714}{RGB}{255,127,14}
\definecolor{forestgreen4416044}{RGB}{44,160,44}
\definecolor{steelblue31119180}{RGB}{31,119,180}

\begin{axis}[
    every axis y label/.style={at={(current axis.north west)},above=0mm},
    legend cell align={left},
    legend style={fill opacity=0.0, draw opacity=1, text opacity=1, draw=none, legend columns=1, column sep=2pt, very thick, anchor=north west, at={(0.0,1.0)}, font=\footnotesize},
    width=6cm, height=4cm,
    tick align=outside,
    tick pos=left,
    xmin=0, xmax=6.2,
    ymin=5, ymax=75,
    xlabel={\# Auxiliary Cameras},
    xtick={0, 1, 2, 4, 6},
    ytick={10, 30, 50, 70},
    ylabel={VLM \textuparrow},
]
\addplot [thick, mark=*, Maroon]
table {%
0 25
1 33.8
2 40.1
4 51.4
6 60.2
};
\addplot [thick, mark=*, ForestGreen]
table {%
0 16.67
1 22.5
2 27.7
4 33.5
6 37.9
};
\addplot [thick, mark=*, NavyBlue]
table {%
0 20.83
1 26.3
2 30.3
4 36.6
6 40.5
};
\draw[gray!65, thin, dashed](axis cs:0,10) -- (axis cs:7,10);
\draw[gray!65, thin, dashed](axis cs:0,30) -- (axis cs:7,30);
\draw[gray!65, thin, dashed](axis cs:0,50) -- (axis cs:7,50);
\draw[gray!65, thin, dashed](axis cs:0,70) -- (axis cs:7,70);
\end{axis}

\end{tikzpicture}}}%
      \caption{\centering \benchname}
      \label{fig:latent_compress}
    \end{subfigure}
    \vspace{-0.2cm}
  \caption{\textbf{Effect of auxiliary camera count.} Adding auxiliary views systematically improves FID and VLM accuracies on consistency and memory metrics across both benchmarks, with diminishing returns once spatial coverage saturates.}
  \label{fig:cam_count}
\end{figure}
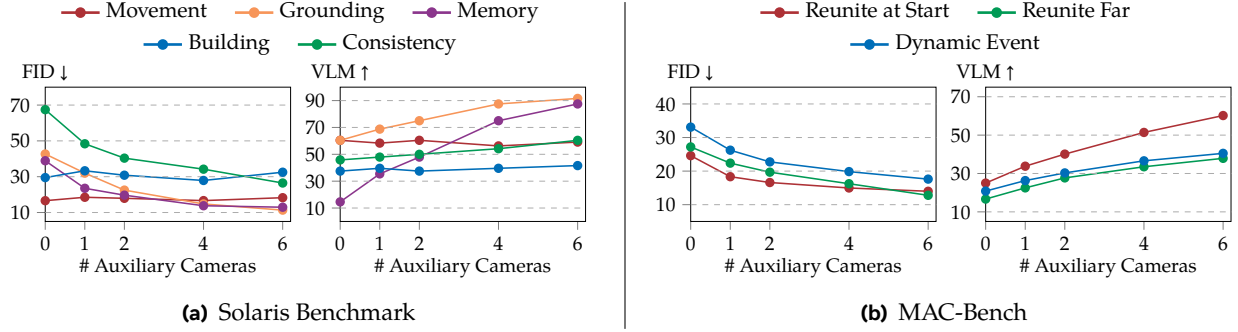

\paragraph{Effect of auxiliary camera count.}
Increasing the number of auxiliary cameras steadily improves every metric relying on persistent world state: grounding, memory, and consistency on Solaris, and all three revisit tasks in \benchname{}, while movement and building performance remain flat (\cref{fig:cam_count}). 
This selective improvement confirms that performance gains stem specifically from expanded spatial visual coverage rather than generic model capacity scaling, which would otherwise uniformly elevate all capabilities. 
Gains naturally plateau once visual coverage saturates.

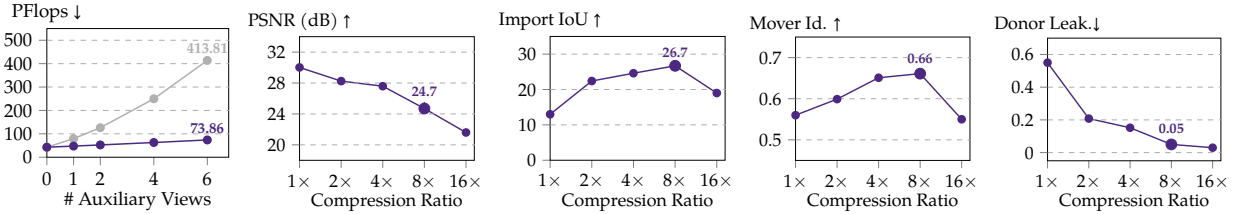
\begin{figure}[ht!]
  \centering
    \begin{subfigure}[t]{0.188\linewidth}
      \centering
        \resizebox{\linewidth}{!}{{\begin{tikzpicture}

\definecolor{c1}{RGB}{176,176,176}
\definecolor{c2}{RGB}{76,40,130}

\begin{axis}[
    every axis y label/.style={at={(current axis.north west)},above=0mm},
    legend cell align={left},
    legend style={fill opacity=0.0, draw opacity=1, text opacity=1, draw=none, legend columns=1, column sep=2pt, very thick, anchor=north west, at={(0.0,1.0)}, font=\footnotesize},
    width=5cm, height=4cm,
    tick align=outside,
    tick pos=left,
    xmin=0, xmax=6.8,
    ymin=0, ymax=550,
    xlabel={\# Auxiliary Views},
    xtick={0, 1, 2, 4, 6},
    ytick={0, 100, 200, 300, 400, 500},
    ylabel={PFlops \textdownarrow},
]
\addplot [thick, mark=*,c1]
table {%
0 42.82
1 79.37
2 126.03
4 249.69
6 413.81
};
\addplot [thick, mark=*, c2]
table {%
0 42.82
1 47.6
2 52.53
4 62.88
6 73.86
};
\draw[-, thick, shorten <=2pt] (axis cs:6,73.86) -- (axis cs:6,73.86) node[pos=1, anchor=south] {\footnotesize \textcolor{c2}{\bf 73.86}};
\draw[-, thick, shorten <=2pt] (axis cs:6,413.81) -- (axis cs:6,413.81) node[pos=1, anchor=south] {\footnotesize \textcolor{c1}{\bf 413.81}};
\draw[gray!65, thin, dashed](axis cs:0,100) -- (axis cs:7,100);
\draw[gray!65, thin, dashed](axis cs:0,200) -- (axis cs:7,200);
\draw[gray!65, thin, dashed](axis cs:0,300) -- (axis cs:7,300);
\draw[gray!65, thin, dashed](axis cs:0,400) -- (axis cs:7,400);
\draw[gray!65, thin, dashed](axis cs:0,500) -- (axis cs:7,500);
\end{axis}

\end{tikzpicture}}}
      \label{fig:cctv_cam}
    \end{subfigure} \hfill
   \begin{subfigure}[t]{0.8\linewidth}
      \centering
      \resizebox{0.25\linewidth}{!}{{\begin{tikzpicture}

\definecolor{c2}{RGB}{76,40,130}

\begin{axis}[
    every axis y label/.style={at={(current axis.north west)},above=0mm},
    legend cell align={left},
    legend style={fill opacity=0.0, draw opacity=1, text opacity=1, draw=none, legend columns=1, column sep=2pt, very thick, anchor=north west, at={(0.0,1.0)}, font=\footnotesize},
    width=5cm, height=4cm,
    tick align=outside,
    tick pos=left,
    xmin=0, xmax=4.2,
    ymin=18, ymax=34,
    xlabel={Compression Ratio},
    xtick={0, 1, 2, 3, 4},
    xticklabels={1$\times$, 2$\times$, 4$\times$, 8$\times$, 16$\times$},
    ytick={20, 24, 28, 32},
    ylabel={PSNR (dB) \textuparrow},
]
\addplot [thick, mark=*, c2]
table {%
0 30.02
1 28.25
2 27.59
3 24.7
4 21.6
};
\addplot [
    only marks,
    mark=*,
    mark size=3pt,
    c2,
] coordinates {
    (3,24.70)
};
\draw[-, thick, shorten <=2pt] (axis cs:3,24.70) -- (axis cs:3,25.5) node[pos=1, anchor=south] {\footnotesize \textcolor{c2}{\bf 24.7}};
\draw[gray!65, thin, dashed](axis cs:0,20) -- (axis cs:5,20);
\draw[gray!65, thin, dashed](axis cs:0,24) -- (axis cs:5,24);
\draw[gray!65, thin, dashed](axis cs:0,28) -- (axis cs:5,28);
\draw[gray!65, thin, dashed](axis cs:0,32) -- (axis cs:5,32);
\end{axis}

\end{tikzpicture}}}%
       \resizebox{0.25\linewidth}{!}{{\begin{tikzpicture}

\definecolor{c2}{RGB}{76,40,130}

\begin{axis}[
    every axis y label/.style={at={(current axis.north west)},above=0mm},
    legend cell align={left},
    legend style={fill opacity=0.0, draw opacity=1, text opacity=1, draw=none, legend columns=1, column sep=2pt, very thick, anchor=north west, at={(0.0,1.0)}, font=\footnotesize},
    width=5cm, height=4cm,
    tick align=outside,
    tick pos=left,
    xmin=0, xmax=4.2,
    ymin=0, ymax=35,
    xlabel={Compression Ratio},
    xtick={0, 1, 2, 3, 4},
    xticklabels={1$\times$, 2$\times$, 4$\times$, 8$\times$, 16$\times$},
    ytick={0, 10, 20, 30},
    ylabel={Import IoU \textuparrow},
]
\addplot [thick, mark=*, c2]
table {%
0 13
1 22.4
2 24.6
3 26.7
4 19
};
\addplot [
    only marks,
    mark=*,
    mark size=3pt,
    c2,
] coordinates {
    (3,26.7)
};
\draw[-, thick, shorten <=2pt] (axis cs:3,26.7) -- (axis cs:3,27.3) node[pos=1, anchor=south] {\footnotesize \textcolor{c2}{\bf 26.7}};
\draw[gray!65, thin, dashed](axis cs:0,10) -- (axis cs:5,10);
\draw[gray!65, thin, dashed](axis cs:0,20) -- (axis cs:5,20);
\draw[gray!65, thin, dashed](axis cs:0,30) -- (axis cs:5,30);
\end{axis}

\end{tikzpicture}}}%
        \resizebox{0.244\linewidth}{!}{{\begin{tikzpicture}

\definecolor{c2}{RGB}{76,40,130}

\begin{axis}[
    every axis y label/.style={at={(current axis.north west)},above=0mm},
    legend cell align={left},
    legend style={fill opacity=0.0, draw opacity=1, text opacity=1, draw=none, legend columns=1, column sep=2pt, very thick, anchor=north west, at={(0.0,1.0)}, font=\footnotesize},
    width=5cm, height=4cm,
    tick align=outside,
    tick pos=left,
    xmin=0, xmax=4.2,
    ymin=0.45, ymax=0.75,
    xlabel={Compression Ratio},
    xtick={0, 1, 2, 3, 4},
    xticklabels={1$\times$, 2$\times$, 4$\times$, 8$\times$, 16$\times$},
    ytick={0.5, 0.6, 0.7},
    ylabel={Mover Id. \textuparrow},
]
\addplot [thick, mark=*, c2]
table {%
0 0.56
1 0.599
2 0.651
3 0.661
4 0.55
};
\addplot [
    only marks,
    mark=*,
    mark size=3pt,
    c2,
] coordinates {
    (3,0.661)
};
\draw[-, thick, shorten <=2pt] (axis cs:3,0.661) -- (axis cs:3,0.67) node[pos=1, anchor=south] {\footnotesize \textcolor{c2}{\bf 0.66}};
\draw[gray!65, thin, dashed](axis cs:0,0.5) -- (axis cs:5,0.5);
\draw[gray!65, thin, dashed](axis cs:0,0.6) -- (axis cs:5,0.6);
\draw[gray!65, thin, dashed](axis cs:0,0.7) -- (axis cs:5,0.7);
\end{axis}

\end{tikzpicture}}}%
       \resizebox{0.25\linewidth}{!}{{\begin{tikzpicture}

\definecolor{c2}{RGB}{76,40,130}

\begin{axis}[
    every axis y label/.style={at={(current axis.north west)},above=0mm},
    legend cell align={left},
    legend style={fill opacity=0.0, draw opacity=1, text opacity=1, draw=none, legend columns=1, column sep=2pt, very thick, anchor=north west, at={(0.0,1.0)}, font=\footnotesize},
    width=5cm, height=4cm,
    tick align=outside,
    tick pos=left,
    xmin=0, xmax=4.2,
    ymin=-0.05, ymax=0.7,
    xlabel={Compression Ratio},
    xtick={0, 1, 2, 3, 4},
    xticklabels={1$\times$, 2$\times$, 4$\times$, 8$\times$, 16$\times$},
    ytick={0.0, 0.2, 0.4, 0.6},
    ylabel={Donor Leak.\textdownarrow},
]
\addplot [thick, mark=*, c2]
table {%
0 0.55
1 0.208
2 0.152
3 0.05
4 0.03
};
\addplot [
    only marks,
    mark=*,
    mark size=3pt,
    c2,
] coordinates {
    (3,0.05)
};
\draw[-, thick, shorten <=2pt] (axis cs:3,0.05) -- (axis cs:3,0.08) node[pos=1, anchor=south] {\footnotesize \textcolor{c2}{\bf 0.05}};
\draw[gray!65, thin, dashed](axis cs:0,0.0) -- (axis cs:5,0.0);
\draw[gray!65, thin, dashed](axis cs:0,0.2) -- (axis cs:5,0.2);
\draw[gray!65, thin, dashed](axis cs:0,0.4) -- (axis cs:5,0.4);
\draw[gray!65, thin, dashed](axis cs:0,0.6) -- (axis cs:5,0.6);
\end{axis}

\end{tikzpicture}}}
      \label{fig:latent_compress}
    \end{subfigure}
    \vspace{-0.5cm}
\caption{\textbf{Effect of $\delta$-VAE compression.} $\delta$-VAE keeps compute nearly flat as auxiliary views are added. Stronger compression trades reconstruction fidelity for disentanglement: donor leakage drops steadily, while mover transfer peaks at the default $8\times$ before degrading.}
  \label{fig:compression_ratio}
\end{figure}

\paragraph{Effect of latent compression.}
In cross-scene transfer experiments (decoding residual codes with a different ``control'' anchor latent), higher compression lowers PSNR but systematically suppresses donor background leakage. 
Crucially, spatial placement precision and identity preservation of the transferred mover (Import IoU and Mover Id.) \textit{improve} up to $8\times$ compression—peaking at $26.7$ and $0.66$, respectively, before degrading at $16\times$ (\cref{fig:compression_ratio}). 
Compression is thus more than an efficiency knob: an informational bottleneck too narrow to encode static background forces the model to allocate capacity to dynamic elements, causing dynamic-static disentanglement to naturally emerge from the constraint itself. We therefore adopt $8\times$ compression by default (\cref{fig:compression_ratio}).

\section{Conclusion}
We introduced \mname{}, a multi-agent video world model that jointly generates egocentric
agent views and fixed-pose auxiliary views as a lightweight proxy for a world-centric state.
These views substantially improve consistency when agents separate and later reunite, and
$\delta$-VAE makes them cheap to model by encoding only changes relative to a static
reference. We hope this simple yet effective design will pave the way for world models that simulate
shared environments beyond what any single agent sees.

\section{Acknowledgement}

The authors would like to thank
Tomas Jakab, Zeren Jiang, Zijian Zhou, Mengmeng Xu, and Wonbong Jang for
their insightful feedback.

\newpage

\bibliographystyle{assets/plainnat}
\bibliography{references,vedaldi_general,vedaldi_specific}
\newpage

\appendix

\section{Dataset Curation Pipelines}
\label{app:data}

\paragraph{Data.}
The three domains are three separate datasets, and we train one model on each. Minecraft and Doom are collected
by us; the real captures are an existing dataset that we re-use in the same two-agent, many-camera arrangement.
In all three, a clip is 241 frames of every view at once, sampled at 20 frames per second (12\,s), $576\times320$
per view in Minecraft and in the captures and $480\times320$ in Doom.

\paragraph{Minecraft.}
The gameplay is collected with the Solaris engine \citep{savva2026solaris}: two Mineflayer bots play a two-player episode
on a Minecraft~1.21 server and each view is recorded from a real Minecraft client, so the frames are the game's
own graphics and every frame carries the bot's ground-truth action. The bots explore by a potential field, walls
pushing them away and unvisited ground pulling them in, which keeps the pair moving over new terrain instead of
circling one spot, and each episode is additionally labeled with one of eleven interaction modes (mining,
building, fighting, chasing, exploring, reuniting and so on). We extend the engine with six fixed auxiliary
cameras. They are placed in a second pass: the episode is played once to establish the terrain, then the six
cameras are sampled at random in the free space around the agents' starting area and the episode is captured with
them. Placement therefore follows the ground the episode covers rather than aiming at one point; measured over
200 episodes, a camera sits a median 4.9 blocks above the agents and 3.0 blocks from the nearest point of their
path, tilted 52 degrees down with a 70 degree field of view, and while a single camera has an agent in its
frustum in about half of the frames, some camera has one in 90\% of them. Every view is recorded at
$1280\times720$ and 20 frames per second. We collect 20{,}000 episodes of outdoor terrain and 11{,}000 of
village worlds, 264 frames each.

\paragraph{Doom.}
Doom episodes are played with Doom \citep{gzdoom} on the 68 maps of Freedoom Phase~1 and Phase~2
\citep{freedoom}, two marines and eight fixed auxiliary cameras per episode, and the marines are driven by the
same potential-field exploration controller as the Minecraft bots. Every view is recorded at $1024\times768$ and
30 frames per second with a 120-degree field of view. The eight cameras sit about 100 map units (3\,m) above eye
height, four tilted 10 degrees down and four 28 degrees, and between them they see the marines almost everywhere
they walk, which is what the maps' corridors and halls allow: a single camera has a marine in its frustum in
about 85\% of the frames. We train on 20{,}000 traversal episodes and hold out a separate set of 4{,}000 arena
episodes, in which the two marines start 128 map units apart facing each other, for evaluation.

\paragraph{Real captures.}
The real sequences come from Replay \citep{shapovalov23replay}, 68 scenes of acted social interaction of about five minutes
each, filmed by twelve temporally synchronized, geometrically and color-calibrated cameras. We use eight of
them in the arrangement the two games are built on: the two head-mounted cameras, which move with their wearers,
are the agent views, and six static cameras are the auxiliary views. Unlike the games this gives real people,
real cameras, and measured poses: every camera comes with a per-frame camera-to-world pose and pinhole
intrinsics; the static cameras hold still to within 5\,mm and a wearer covers 3 to 4\,m in a clip. We hold out
45 clips of 241 frames, drawn from 35 of the scenes, and train on the rest of the dataset.

\paragraph{\benchname{}.}
\Cref{tab:macb_spec} lists the task specifications.
In all tasks, the agents begin mutually visible and lose sight of each other mid-episode.
In \emph{Dynamic Event}, the scripted event is placed at the midpoint between the agents and
is one of sprouting crops, a standing creeper, or a plank cube that ignites at frame $20$.
Doom uses the two reunite tasks on six Freedoom maps with $1047$-unit walks ($32$ units per
block), with the partner $84^\circ$ off-axis at the midpoint.
We use $32$ episodes per task in Minecraft and $100$ per task in Doom.
For VLM accuracy, Claude Opus 5 is given the generated frames and answers the task query for each agent;

An episode counts as correct only if the answers for both agents match the ground truth.
FID is computed between generated and ground-truth agent frames.
During evaluation, accuracy is averaged over 5 independent runs of 32 episodes; episodes with failed generation or unparsable judge output are excluded.

\begin{table}[ht!]
\centering
\small
\setlength{\tabcolsep}{4pt}
\resizebox{\linewidth}{!}{%
\begin{tabular}{@{}lllll@{}}
\toprule
Task & Start & Mid-episode & End & Query \\
\midrule
Reunite at Start & $13$ blocks apart, facing
  & each wanders $\le 9$ blocks
  & exact start poses
  & Is the partner visible? \\
Reunite Far & $17$ blocks apart
  & $\sim\!59$ blocks each, converging; partner $86^\circ$ off-axis
  & $2.9$ blocks apart, facing
  & Is the partner visible? \\
Dynamic Event & $13$ blocks apart, event at midpoint
  & separate to $40$ blocks by frame $76$
  & $4.3$ blocks apart
  & What happened to the event? \\
\bottomrule
\end{tabular}}
\caption{\textbf{\benchname{} task specifications in Minecraft.} Mid-episode angles are
medians over episodes.}
\label{tab:macb_spec}
\end{table}

\section{Auxiliary Camera Pose Sampling}%
\label{app:pose_sampling}

As described in \cref{sec:model_design}, \mname requires auxiliary camera poses $\mathbf{c}$ in \cref{eq:joint_world_model} but no auxiliary images at inference time.
During training, auxiliary camera poses and their corresponding videos are provided by the dataset.
At inference time, we sample the poses as follows.

\paragraph{Scene-Independent Pose Prior.}

We define a canonical coordinate system whose origin is the initial position of the first agent.
Candidate auxiliary camera poses are sampled relative to this coordinate system from a scene-independent camera-pose prior.
The horizontal offset from the origin, camera height, yaw, and pitch are modeled as Gaussian variables whose parameters are estimated from the training camera poses.
Camera height is constrained to be non-negative, and roll is fixed to zero.

\paragraph{Coverage-Based Selection.}

We sample a candidate set of poses $\mathcal{C}$ from this prior and greedily select $K$ cameras to maximize geometric field-of-view coverage.
Let $\Omega$ denote a bounded discretized volume in the canonical coordinate system, and let $\mathcal{F}(\mathbf{c})\subseteq\Omega$ denote the cells geometrically covered by the camera frustum associated with pose $\mathbf{c}$.
The computation of $\mathcal{F}(\mathbf{c})$ depends only on camera geometry and does not use scene occupancy or occlusion information.
Let $\mathcal{R}_k$ denote the region covered by the first $k$ selected cameras, with $\mathcal{R}_0=\emptyset$.
We greedily select
\begin{align}
\mathbf{c}_{k+1}
&=
\arg\max_{\mathbf{c}\in\mathcal{C}}
\left|
\mathcal{F}(\mathbf{c})\setminus\mathcal{R}_k
\right|,
\qquad
\mathcal{R}_{k+1}
=
\mathcal{R}_k\cup\mathcal{F}(\mathbf{c}_{k+1}),
\end{align}
until $K$ cameras are selected.
Since both candidate sampling and coverage computation depend only on the canonical coordinate system and camera geometry, this procedure requires no additional scene-specific visual or geometric input.

\section{More Ablations}
\label{app:more_ablations}
\begin{table*}[ht]
\centering
\footnotesize
\setlength{\tabcolsep}{5.0pt}
\renewcommand{\arraystretch}{1.06}

\begin{tabular}{@{}l c cc cc cc@{}}
\toprule
& \multicolumn{1}{c}{\textbf{$\delta$-VAE}}
& \multicolumn{6}{c}{\textbf{Minecraft Multi-agent Consistency Benchmark}} \\
\cmidrule(lr){2-2}
\cmidrule(lr){3-8}

& \textbf{PSNR$\uparrow$}
& \multicolumn{2}{c}{\textbf{Reunite at Start}}
& \multicolumn{2}{c}{\textbf{Reunite Far}}
& \multicolumn{2}{c}{\textbf{Dynamic Event}} \\
\cmidrule(lr){3-4}
\cmidrule(lr){5-6}
\cmidrule(lr){7-8}

\textbf{Variant}
& \textbf{dB}
& \textbf{VLM$\uparrow$} & \textbf{FID$\downarrow$}
& \textbf{VLM$\uparrow$} & \textbf{FID$\downarrow$}
& \textbf{VLM$\uparrow$} & \textbf{FID$\downarrow$} \\
\midrule

w/o $\mathcal{L}_{\mathrm{align}}$
& 25.1
& 56.9 & 13.47
& 33.4 & 12.26
& 38.1 & 17.08 \\

\rowcolor{mnameblue}
Full $\delta$-VAE
& 24.7
& 60.2 & 13.99
& 37.9 & 12.79
& 40.5 & 17.61 \\

\bottomrule
\end{tabular}%

\caption{\textbf{Ablation of the feature alignment loss.}
We evaluate its effect on $\delta$-VAE reconstruction quality and downstream
world consistency in Minecraft.}
\label{tab:feature_alignment_ablation}
\end{table*}

\paragraph{Feature alignment loss.}
\Cref{tab:feature_alignment_ablation} ablates $\mathcal{L}_{\text{align}}$.
Removing it slightly improves $\delta$-VAE reconstruction ($25.1$ vs.\ $24.7$\,dB PSNR) and
FID, but lowers VLM accuracy on all three \benchname{} tasks, by $2.4$--$4.5$ points.
Without alignment, the residual code spends its capacity on reconstructing the whole frame;
with it, the code is pushed towards dynamic regions, which is what the world model needs to
keep revisited scenes consistent.
Reconstruction fidelity is thus not a sufficient proxy for how useful a compressed
representation is downstream, in line with the compression trend in
\cref{sec:exp_ablation}.

\section{Implementation Details}
\label{app:impl}
By applying $2\times2$ spatial ($s=2$) and $2\times$ temporal ($\tau = 2$) compression rates, the $\delta$-VAE reduces the auxiliary view representation from $864$ tokens per frame (at $576\times384$ resolution) to just $108$ tokens, achieving an $8\times$ overall bandwidth reduction.
\mname jointly generates both agents' egocentric streams and $K$ auxiliary views ($K{=}6$ by default for Minecraft and $K{=}8$ for Doom) over a 241-frame horizon, conditioned on the initial frame of agent views and both agents' action sequences.
Built upon Wan2.2-TI2V-5B~\citep{wan2025}, the model is fully fine-tuned per domain for 20k steps using Adam with a learning rate of $10^{-4}$ and a batch size of 64 across 64 NVIDIA GB300 GPUs.
Training incorporates diffusion forcing and causal masking over a context window of 12 latent frames (48 RGB frames), with autoregressive rollout used during inference. 

For training the $\delta$-VAE, we use Adam with a learning rate of $10^{-4}$ and we set $\lambda_{\mathrm{rec}}=1$, $\lambda_{\mathrm{KL}}=10^{-5}$, and $\lambda_{\mathrm{align}}=10^{-3}$.

\section{Limitations and Future Work}
\mname's benefits depend on auxiliary coverage. Regions far from the sampled cameras, as in \textit{Reunite Far}, remain harder to keep consistent. Short-horizon egocentric capabilities such as movement do not benefit from auxiliary views. Aggressive residual compression also trades some fidelity and, in fast-changing scenes such as Doom, consistency for efficiency. Promising directions include adaptive or learned camera placement that follows agents into unexplored regions, auxiliary views that are re-anchored for unbounded worlds, and scaling to larger agent populations. More broadly, we hope this work encourages treating the \textit{output space} of world models, meaning what they choose to render and not only what they condition on, as a first-class design axis for building coherent, shared simulated worlds.

\section{Prompts}
\label{app:prompts}

\ifdefined\promptquote\else   %
\newenvironment{promptquote}
  {\begin{quote}\small\ttfamily\raggedright}
  {\end{quote}}
\fi

\subsection{The judge}
\label{app:prompts_judge}

Every question in both benchmarks is asked of the same judge, Claude Opus~5, in one turn: the
frame or the pair of frames first, then the question, under the system prompt \emph{``You are a
helpful assistant that evaluates Minecraft screenshots.''} Thinking is adaptive at low effort.
The answer is reduced to its first non-empty line, lowercased and stripped of quotes and
trailing punctuation, and validated by exact match against the expected word, so every prompt
ends by naming the words it accepts. An episode counts as correct only if every question about
it is answered correctly.

\paragraph{Revisiting benchmark.}
The two reunite tasks ask only about the partner, in one sentence:
\begin{promptquote}
Here is a Minecraft/Doom screenshot. Is there another player visible on-screen? Answer with a single
word: "yes", "no".
\end{promptquote}
It is put to each agent's own view at four moments read off the recorded trajectory: the first
generated frame, the frame of maximum separation, the midpoint of the way back, and the
rendezvous, which makes eight questions per episode. The judge sees one frame and this sentence,
with no description of the task and no earlier frame, so nothing but the frame can decide the
answer; the expected answer comes from the poses, the partner counting as visible when it lies
well inside the view cone and within 20 blocks and as not visible when it lies outside the cone
or beyond 35 blocks, with the judge's own answer on the ground-truth frame standing in for the
band between the two. The dynamic
events add two more question shapes. The outcome question names the states the event can be in,
one set per event kind; for the creeper,
\begin{promptquote}
Here is a Minecraft screenshot. What is visible where the creeper stood: a creeper still
standing, a crater or hole blasted into the ground, or flat undamaged ground? Answer with a
single word from "creeper", "crater", "flat".
\end{promptquote}
with ``sprouts''/``wheat''/``none'' for the growing crops, ``intact''/``burning''/``partly
burnt''/``gone'' for the burning plank cube. The progression question shows the first
and the return frame together:
\begin{promptquote}
Here are two Minecraft screenshots from roughly the same viewpoint, the first taken earlier and
the second later. In this scene \{process\}. Compared with the first image, is that process
further along in the second image (progressed), at the same stage (same), or at an earlier stage
than in the first image (reverted)? Answer with a single word: "progressed", "same", "reverted".
\end{promptquote}
where \{process\} is a phrase for the event, for instance \emph{``a creeper explodes, disappears,
and leaves a crater blasted into the ground''}. Each event kind also has a presence question
asked at the first generated frame (\emph{``Is there a creeper (green four-legged hostile mob)
visible on-screen?''} and its equivalents).

\paragraph{One sentence we add.}
Claude reads the viewer's own first-person arm in the corner of a sharp frame as another player.
Where that matters, we append, to every prompt and identically in both passes,
\begin{promptquote}
The large blocky shape in the bottom-right corner is the viewer's own first-person arm and held
item, not another player.
\end{promptquote}
The prompt actually used is recorded in every result file.

\subsection{Action prompts}
\label{app:prompts_action}

Actions reach the model as language. Each of the 28 entries of the action vocabulary is written
as one English sentence, encoded once by the text encoder, and stored as a table of token
sequences; at training and at inference the sentence of the frame's action is looked up and its
tokens are used as the frame's conditioning context. \Cref{tab:action_prompts} lists the
sentences.

Per frame exactly one entry is chosen. The keyboard flags score one when pressed; the mouse
contributes through its yaw and pitch, whose sign picks turning left or right and looking up or
down and whose magnitude is clipped to one; if nothing scores above zero, the frame is idle. The
highest score wins, so a frame that both walks and turns is conditioned on the stronger of the
two. The conditioning frame's tokens are zeroed, since it is given rather than generated, and
during training the whole action context is dropped with a fixed probability so that the model
can be sampled with classifier-free guidance on the actions.

\begin{table}[t]
\centering\small
\begin{tabular}{@{}ll@{}}
\toprule
Entry & Sentence \\
\midrule
\texttt{forward}          & The player is walking forward \\
\texttt{left}             & The player is strafing to the left \\
\texttt{back}             & The player is walking backward \\
\texttt{right}            & The player is strafing to the right \\
\texttt{jump}             & The player is jumping into the air \\
\texttt{sneak}            & The player is sneaking and crouching down \\
\texttt{sprint}           & The player is sprinting and running fast \\
\texttt{drop}             & The player is dropping the held item \\
\texttt{inv}              & The player is opening the inventory \\
\texttt{esc}              & The player is pressing escape to pause \\
\texttt{tab}              & The player is switching the active hotbar tab \\
\texttt{hot1}--\texttt{hot9} & The player is selecting item slot one \ldots\ nine in the hotbar \\
\texttt{swap\_hands}      & The player is swapping the items in both hands \\
\texttt{attack}           & The player is attacking and hitting with their weapon \\
\texttt{use}              & The player is using or interacting with an object \\
\texttt{mouse\_turn\_left}  & The player is turning the camera left \\
\texttt{mouse\_turn\_right} & The player is turning the camera right \\
\texttt{mouse\_look\_up}    & The player is looking up \\
\texttt{mouse\_look\_down}  & The player is looking down \\
\texttt{idle}             & The player is standing still and doing nothing \\
\bottomrule
\end{tabular}
\caption{The action vocabulary, one sentence per entry. The nine hotbar entries differ only in
the number they name. \texttt{drop}, \texttt{inv}, \texttt{esc}, \texttt{tab} and
\texttt{swap\_hands} are never pressed in our episodes and are kept only so that the vocabulary
matches the game's input.}
\label{tab:action_prompts}
\end{table}

\section{More Visualizations}
\label{appendix:more_vis}

\begin{figure}[ht!]
  \centering
  \includegraphics[width=\textwidth]{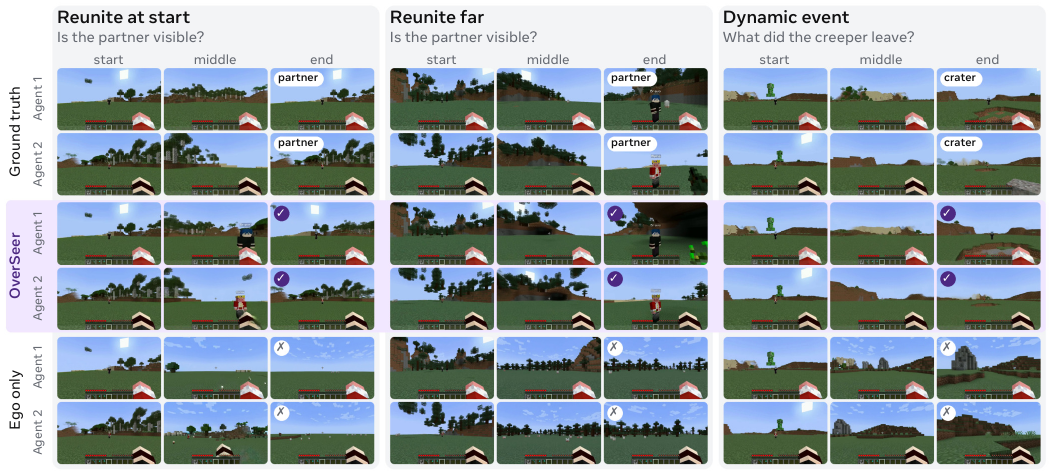}\\[4pt]
  \includegraphics[width=\textwidth]{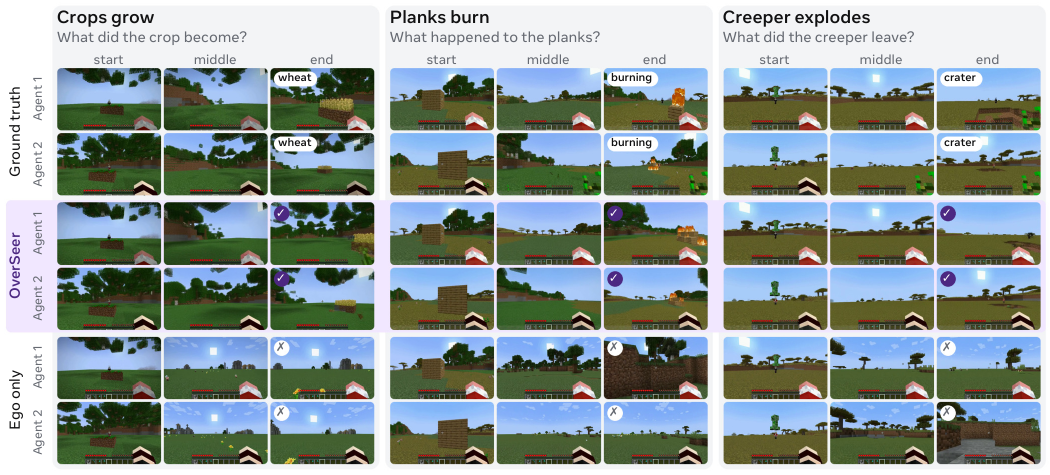}
  \caption{\textbf{Additional \benchname{} results in Minecraft.} Top: the three benchmark
  tasks. Bottom: three further dynamic events. \mname{} recovers the partner or the evolved
  event state where the ground truth does, while the egocentric-only variant does not.}
  \label{fig:campaign_modes}
\end{figure}

\begin{figure}[ht!]
  \centering
  \includegraphics[width=\textwidth]{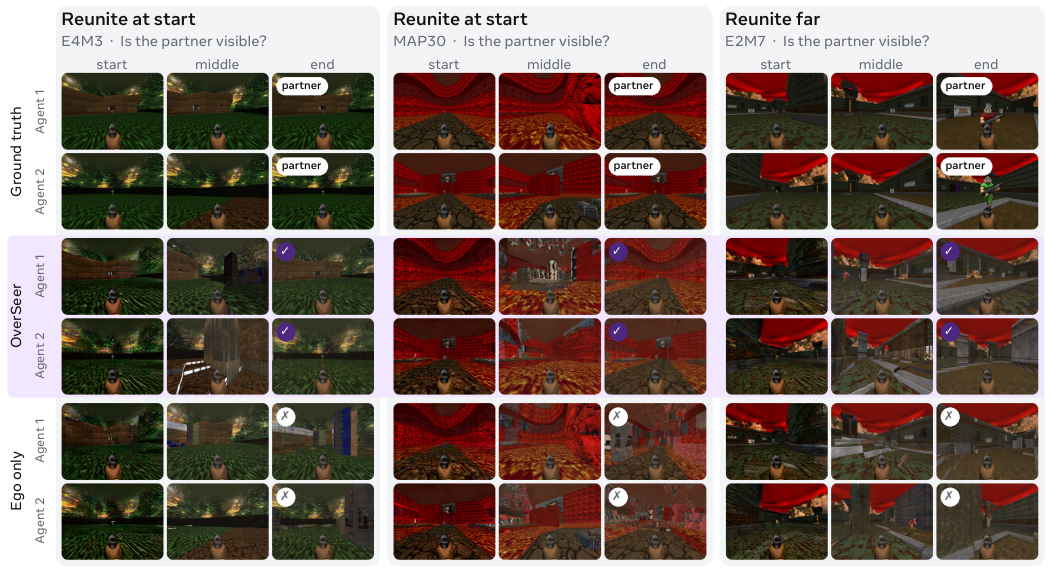}
  \caption{\textbf{\benchname{} results in Doom.} Reunite tasks on three Freedoom maps.
  \mname{} restores the partner after the agents separate; the egocentric-only variant does
  not.}
  \label{fig:doom_2ego_ablation}
\end{figure}

\begin{figure}[ht!]
  \centering
  \includegraphics[width=\textwidth]{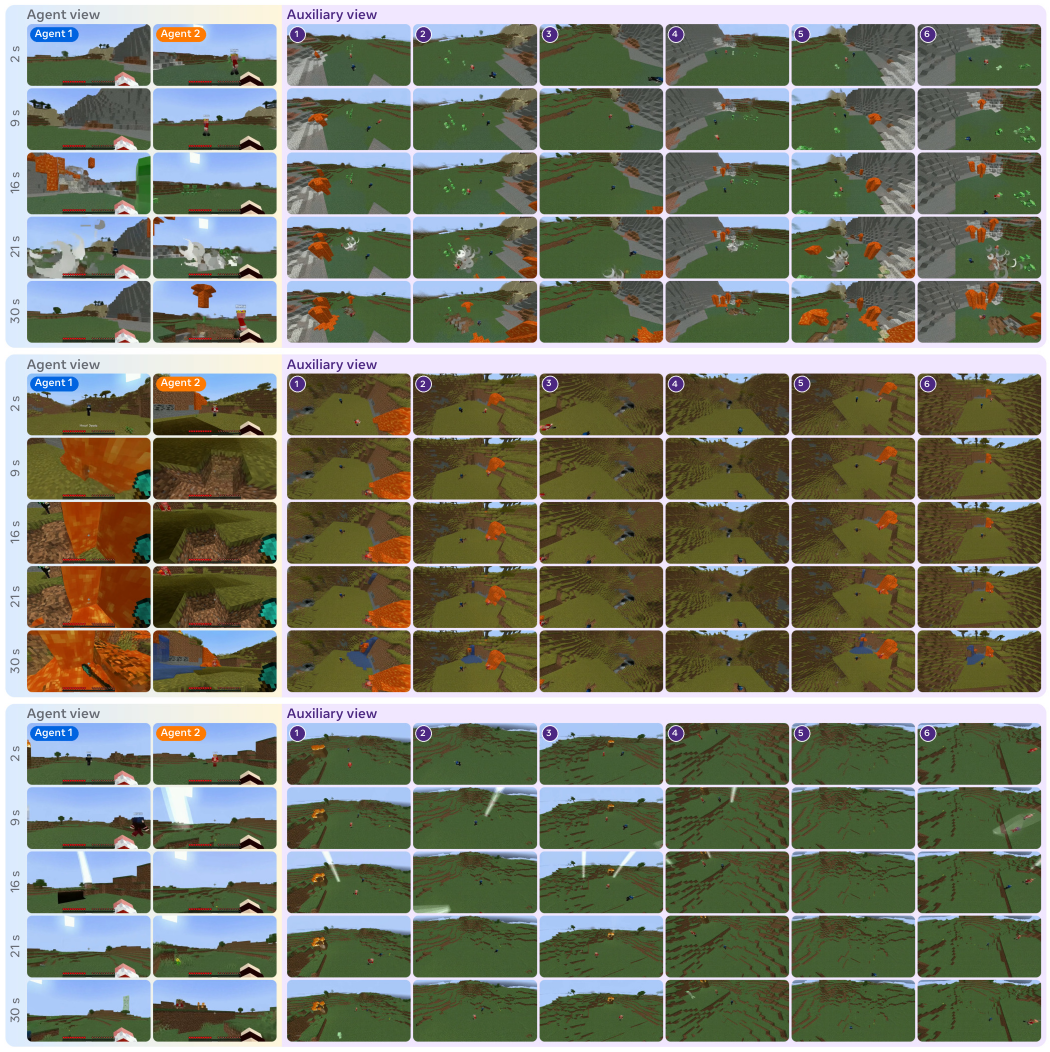}
  \caption{\textbf{Long rollouts in Minecraft.} Each block shows both agent views and all
  six auxiliary views over $30$\,s.}
  \label{fig:minecraft_samples}
\end{figure}

\begin{figure}[ht!]
  \centering
  \includegraphics[width=\textwidth]{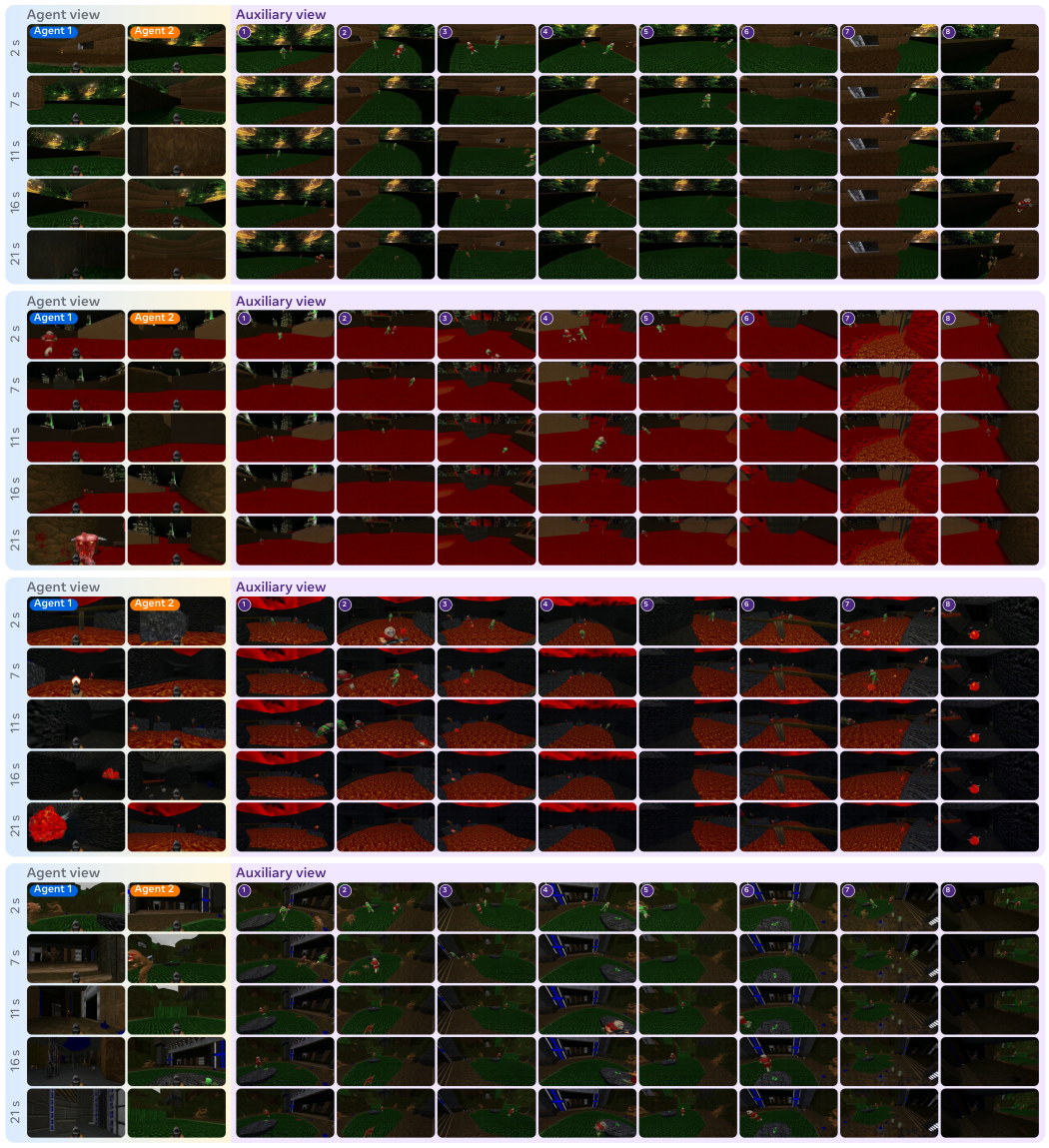}
  \caption{\textbf{Long rollouts in Doom.} Each block shows both agent views and all eight
  auxiliary views over $21$\,s.}
  \label{fig:doom_samples}
\end{figure}

\begin{figure}[ht!]
  \centering
  \includegraphics[width=\textwidth]{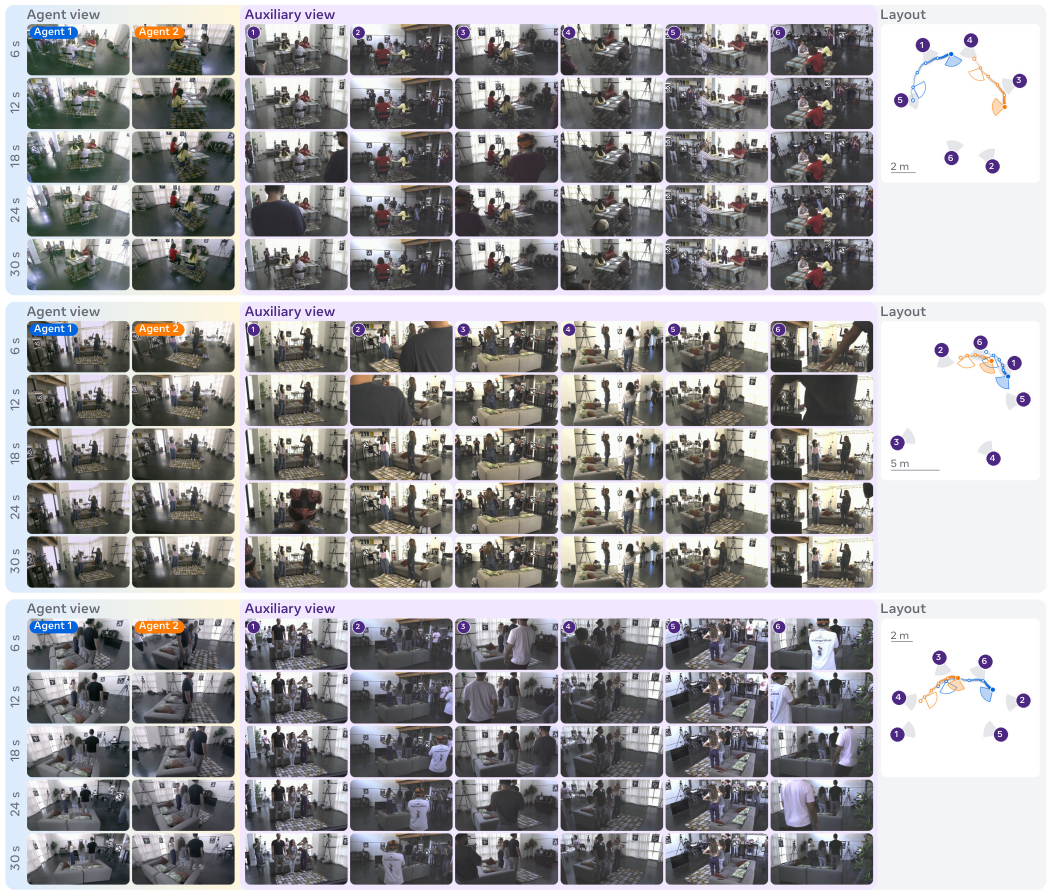}
  \caption{\textbf{Long rollouts on real-world multi-camera captures.} Left: both agent
  views and all six auxiliary views over $30$\,s. Right: top-down layout of the auxiliary
  cameras and agent trajectories.}
  \label{fig:capture_samples}
\end{figure}

\Cref{fig:campaign_modes,fig:doom_2ego_ablation} show additional \benchname{}
episodes in Minecraft and Doom, comparing \mname{} with an egocentric-only variant trained
without auxiliary views.
\Cref{fig:minecraft_samples,fig:doom_samples,fig:capture_samples} show long rollouts
in Minecraft, Doom and real-world captures, with both agent views and all auxiliary views;
for real captures we also show a top-down layout of the cameras and agent trajectories.
Videos are available on our website.

\end{document}